%% file: draft.tex
\documentclass[sigconf,authorversion=true]{acmart}

\usepackage{booktabs} % For formal tables

\usepackage{graphicx}
\usepackage{subfig}
\usepackage[linesnumbered, vlined, ruled]{algorithm2e}
\def\vector#1{\mbox{\boldmath $#1$}}

\setcopyright{rightsretained}
\acmDOI{10.1145/nnnnnnn.nnnnnnn}

\acmISBN{978-x-xxxx-xxxx-x/YY/MM}

\copyrightyear{2019}
\acmYear{2019}
\setcopyright{acmcopyright}
\acmConference[GECCO '19]{Genetic and Evolutionary Computation Conference}{July 13--17, 2019}{Prague, Czech Republic}
\acmBooktitle{Genetic and Evolutionary Computation Conference (GECCO '19), July 13--17, 2019, Prague, Czech Republic}
\acmPrice{15.00}
\acmDOI{10.1145/3321707.3321754}
\acmISBN{978-1-4503-6111-8/19/07}

\begin{document}

\title[Non-elitist Evolutionary Multi-objective Optimizers Revisited]{Non-elitist Evolutionary Multi-objective Optimizers Revisited}

%Using an Unbounded External Archive}

%% \author{\hspace{1em}}
%% %\authornote{Dr.~Trovato insisted his name be first.}
%% %\orcid{1234-5678-9012}
%% \affiliation{%
%%   \institution{\hspace{1em}}
%%   \streetaddress{\hspace{1em}}
%%   \city{\hspace{1em}} 
%% %  \state{\hspace{1em}} 
%%   \postcode{\hspace{1em}}
%% }

%% \author{\hspace{1em}}
%% %\authornote{Dr.~Trovato insisted his name be first.}
%% %\orcid{1234-5678-9012}
%% \affiliation{%
%%   \institution{\hspace{1em}}
%%   \streetaddress{\hspace{1em}}
%% %  \city{\hspace{1em}} 
%% %  \state{\hspace{1em}} 
%% %  \postcode{\hspace{1em}}
%% }
%\email{\,}

%% \author{\hspace{1em}}
%% %\authornote{Dr.~Trovato insisted his name be first.}
%% %\orcid{1234-5678-9012}
%% \affiliation{%
%%   \institution{\hspace{1em}}
%%   \streetaddress{\hspace{1em}}
%%   \city{\hspace{1em}} 
%% %  \state{\hspace{1em}} 
%%   \postcode{\hspace{1em}}
%% }
%% %\email{a@a}

\author{Ryoji Tanabe}
\affiliation{%
  \institution{
Shenzhen Key Laboratory of Computational Intelligence,\\
University Key Laboratory of Evolving Intelligent Systems of Guangdong Province,\\
Department of Computer Science and Engineering,\\
Southern University of Science and Technology}
}
\email{rt.ryoji.tanabe@gmail.com}

%\author{Hisao Ishibuchi (Corresponding author)}
\author{Hisao Ishibuchi}
\authornote{Corresponding author: Hisao Ishibuchi}
\affiliation{%
  \institution{
Shenzhen Key Laboratory of Computational Intelligence,\\
University Key Laboratory of Evolving Intelligent Systems of Guangdong Province,\\
Department of Computer Science and Engineering,\\
Southern University of Science and Technology}
}
\email{hisao@sustech.edu.cn}

\input{abstract.tex}
%% \begin{abstract}
%% This paper provides a sample of a \LaTeX\ document which conforms,
%% somewhat loosely, to the formatting guidelines for
%% ACM SIG Proceedings.\footnote{This is an abstract footnote}
%% \end{abstract}

%
% The code below should be generated by the tool at
% http://dl.acm.org/ccs.cfm
% Please copy and paste the code instead of the example below. 
%



%real-coded crossover methods
\maketitle

\input{introduction.tex}

\input{preliminaries.tex}

\input{experimental_settings.tex}

\input{experimental_results.tex}

%% \input{discussion.tex}
\input{conclusion.tex}

 \section*{Acknowledgments}

%% Acknowledgement
%% Acknowledgments

This work was supported by National Natural Science Foundation of China (Grant No. 61876075), the Program for Guangdong Introducing Innovative and Enterpreneurial Teams (Grant No. 2017ZT07X386), Shenzhen Peacock Plan (Grant No. KQTD2016112514355531), the Science and Technology Innovation Committee Foundation of Shenzhen (Grant No. ZDSYS201703031748284), and the Program for University Key Laboratory of Guangdong Province (Grant No. 2017KSYS008).

\bibliographystyle{ACM-Reference-Format}
\bibliography{reference} 

\end{document}

%% file: abstract.tex
\begin{abstract}
  
Since around 2000, it has been considered that elitist evolutionary multi-objective optimization algorithms (EMOAs) always outperform non-elitist EMOAs.
This paper revisits the performance of non-elitist EMOAs for bi-objective continuous optimization when using an unbounded external archive.
This paper examines the performance of EMOAs with two elitist and one non-elitist environmental selections.
The performance of EMOAs is evaluated on the bi-objective BBOB problem suite provided by the COCO platform.
In contrast to conventional wisdom, results show that non-elitist EMOAs with particular crossover methods perform significantly well on the bi-objective BBOB problems with many decision variables when using the unbounded external archive.
This paper also analyzes the properties of the non-elitist selection.

\end{abstract}

%% file: introduction.tex
\section{Introduction}
\label{sec:introduction}

%f_1 (\vector{x}), ..., f_M

%Multi-objective optimization problems (MOPs) appear in real-world applications.
% $\vector{x}$   $\vector{x}^{\rm final}$ 

Since no solution can simultaneously minimize multiple conflicting objective functions in general, the ultimate goal of multi-objective optimization problems (MOPs) is to find a Pareto optimal solution preferred by a decision maker \cite{Miettinen98}.
When the decision maker's preference information is unavailable a priori, an ``a posteriori'' decision making is performed.
The decision maker selects the final solution from a solution set that approximates the Pareto front in the objective space.

%Since EMOAs are population-based optimizers, they are likely to find a set of solutions in a single run.

%An evolutionary multi-objective optimization algorithm (EMOA) is frequently used for the ``a posteriori'' decision making \cite{Deb01}.

An evolutionary multi-objective optimization algorithm (EMOA) is frequently used to find an approximation of the Pareto front
for the ``a posteriori'' decision making \cite{Deb01}.
A number of EMOAs have been proposed in the literature.
Classical EMOAs include VEGA \cite{Schaffer85}, MOGA \cite{FonsecaF93}, and NSGA \cite{SrinivasD94} proposed in the 1980s and 1990s.
They are non-elitist EMOAs, which do not have a mechanism to maintain non-dominated solutions in the population.
Some elitist EMOAs (e.g., SPEA \cite{ZitzlerT99}, SPEA2 \cite{ZitzlerLT01}, and NSGA-II \cite{DebAPM02}) have been proposed in the early 2000s.
Elitist EMOAs explicitly keep non-dominated solutions found during the search process.

%Unlike classical non-elitist EMOAs, 

Some EMOAs store non-dominated solutions found so far in an unbounded or bounded external archive independently from the population.
For example, MOGLS \cite{IshibuchiM98} proposed in the mid-1990s does not maintain elite solutions in the population but stores all non-dominated solutions found so far in the unbounded external archive.
$\epsilon$-MOEA \cite{DebMM05} stores non-dominated solutions in the population and $\epsilon$-nondominated solutions in the unbounded external archive.
PESA \cite{CorneKO00} uses the non-elitist population and the elitist bounded external archive. % that stores non-dominated solutions.
%The aim of using the external archive in these EMOAs (e.g., MOGLS, $\epsilon$-MOEA, and PESA) is twofold.
The external archive in these EMOAs (e.g., MOGLS, $\epsilon$-MOEA, and PESA) plays two roles.
The first role is to provide non-dominated solutions found so far to the decision maker.
The performance of these types of EMOAs is also evaluated based on solutions in the external archive, rather than the population.
The second role is to perform an elitist search. % using non-dominated solutions found so far.
For example, parents for mating are selected from the external archive in PESA.
Some elitist individuals in the external archive can enter the population in MOGLS.
Since these types of EMOAs explicitly exploit elitist solutions as explained above, they can be categorized into elitist EMOAs.

Apart from algorithm development, the external archive has been used only for the first role (e.g., \cite{FonsecaF93,Lopez-IbanezKL11,BringmannFK14,BrockhoffTH15,WessingPBR17}).
As pointed out in \cite{BringmannFK14}, good potential solutions found so far are likely to be discarded from the population.
The external archive that stores all non-dominated solutions independently from EMOAs can address this issue.
The external archive for the first role can be incorporated into all EMOAs without any changes in their algorithmic behavior.
The external archive is useful for real-world problems where the evaluation of each solution is expensive, i.e., the total number of examined solutions is limited, and the archive maintenance cost is relatively small in comparison with the solution evaluation cost.
If the decision maker wants to examine a small number of non-dominated solutions, solution selection methods are available such as hypervolume indicator-based selection methods (e.g., \cite{BringmannFK14}) and distance-based selection methods (e.g., \cite{SinghBR19}).

This paper revisits non-elitist EMOAs with the unbounded external archive only for the first role (performance evaluation).
When the performance of EMOAs is evaluated based on solutions in the external archive as in \cite{Lopez-IbanezKL11,BringmannFK14,BrockhoffTH15,WessingPBR17}, the role of EMOAs is only to find non-dominated solutions with high quality.
Thus, EMOAs do not need to maintain non-dominated solutions found so far in the current population with the population size $\mu$.
We investigate three environmental selections: best-all (BA), best-family (BF), and best-children (BC).
While BA and BF are elitist selections, BC is a non-elitist selection.
Although BA is a traditional $(\mu+\lambda)$-selection, BF and BC restrict a selection only among $k$ parents and $\lambda$ children.
Thus, $\mu-k$ non-parents do not directly participate in the selection process in BF and BC unlike traditional $(\mu+\lambda)$- and $(\mu,\lambda)$-selections.
In BC, all $k$ parents are removed from the population regardless of their quality.
Then, the top-ranked $k$ out of $\lambda$ children enter the population.
Subsection \ref{sec:env_selection} explains BA, BF, and BC in detail.
%Classical non-elitist EMOAs (e.g., MOGA) and BC are similar in that parents are deleted from the population with no comparison.
%
% In contrast to the above-mentioned elitist EMOAs, this type of non-elitist EMOAs has not been studied well in the 2000's.
%Unlike elitist EMOAs (e.g., MOGLS, $\epsilon$-MOEA, and PESA), the three selections (BA, BF, and BC) do not use non-dominated solutions in the external archive for the search.
%
We examine the performance of EMOAs with the three selections on the bi-objective BBOB problem suite \cite{TusarBHA16}.
We use five crossover methods and four ranking methods in representative EMOAs.

%We use five crossover methods (SBX \cite{DebA95}, BLX \cite{EshelmanS92}, PCX \cite{DebAJ02}, SPX \cite{TsutsuiYH99}, and REX \cite{AkimotoSOK09}) and ranking methods in four EMOAs (NSGA-II \cite{DebAPM02}, SMS-EMOA \cite{BeumeNE07}, SPEA2 \cite{ZitzlerLT01}, and IBEA \cite{ZitzlerK04}).

%using five crossover methods (SPX \cite{TsutsuiYH99} and REX \cite{AkimotoSOK09}) and ranking methods in four EMOAs (NSGA-II, SPEA2, SMS-EMOA, and IBEA) 

Our contributions in this paper are at least threefold:

\begin{itemize}
\item We demonstrate that the non-elitist BC selection performs significantly well on the bi-objective BBOB problems with many decision variables when using the unbounded external archive.
  Although most EMOAs proposed in the 2000s are elitist EMOAs, our results indicate that efficient non-elitist EMOAs could be designed.
 Thus, our results significantly expand the design possibility of EMOAs.
  %Our result is significantly inconsistent with the conventional wisdom ``elitist EMOAs always outperform non-elitist EMOAs''.
\item We demonstrate that restricted replacements in BF and BC are suitable for crossover methods with the preservation of statistics \cite{KitaOK98} (e.g., the property where the covariance matrix of children is the same as that of the parents) such as SPX \cite{TsutsuiYH99} and REX \cite{AkimotoSOK09}.
\item We discuss why the simple BA selection performs worse than the restricted BF and BC selections. We also analyze the properties of the non-elitist BC selection.
 %\item We demonstrate that restricted replacements in BF and BC are suitable for crossover methods with the preservation of statistics \cite{KitaOK98} (i.e., the property of crossover where the covariance matrix of children is the same as that of the parents). EMOAs with BF and BC perform significantly well when using SPX and REX.
  %\item We
\end{itemize}

The rest of this paper is organized as follows.
Section \ref{sec:preliminaries} provides some preliminaries of this paper, including the definition of MOPs, the five crossover methods, and the three environmental selections.
Section \ref{sec:experimental_settings} describes the experimental setup. 
Section \ref{sec:experimental_results} examines the performance of the three environmental selections.
Section \ref{sec:conclusion} concludes this paper with discussions on future research directions.

%% file: preliminaries.tex
\begin{figure*}[t]
\newcommand{\widthvar}{0.2}
  \begin{center} 
    %
    %% \subfloat[AGR]{\includegraphics[width=\widthvar\textwidth]{graph/tmp/pprldmany_05D_noiselessall.pdf}}
    \subfloat[SBX]{\includegraphics[width=\widthvar\textwidth]{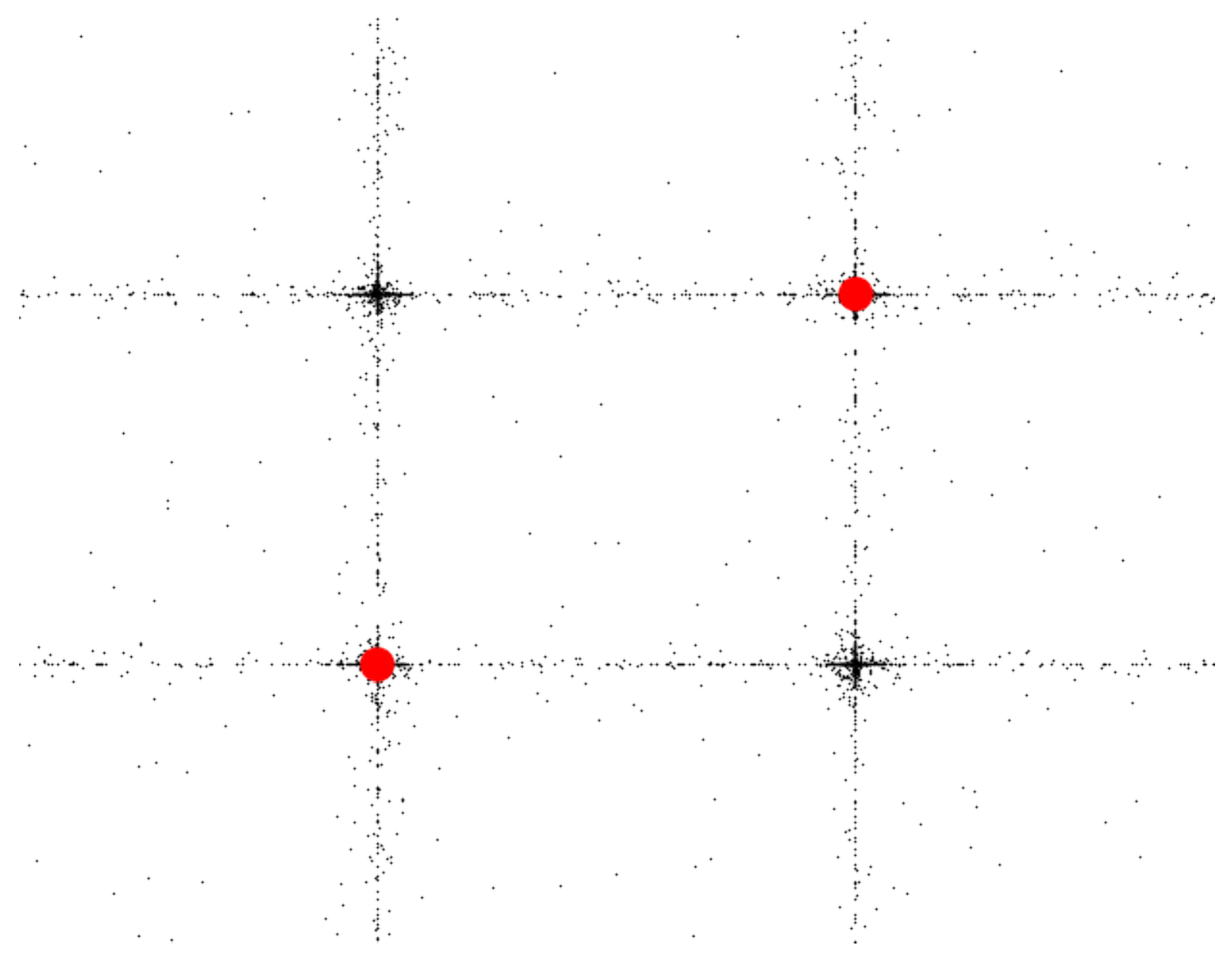}}
    \subfloat[BLX]{\includegraphics[width=\widthvar\textwidth]{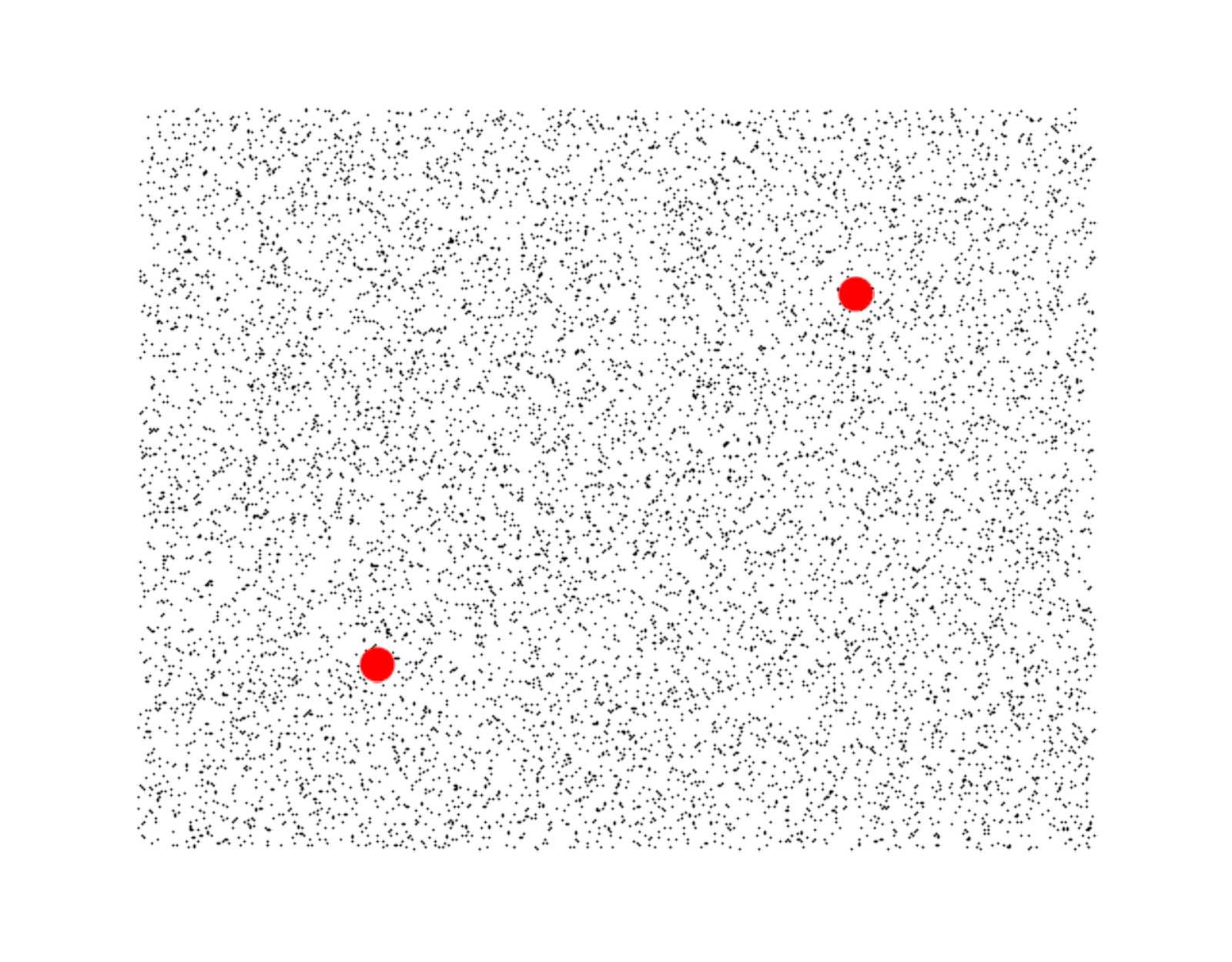}}
    \subfloat[PCX]{\includegraphics[width=\widthvar\textwidth]{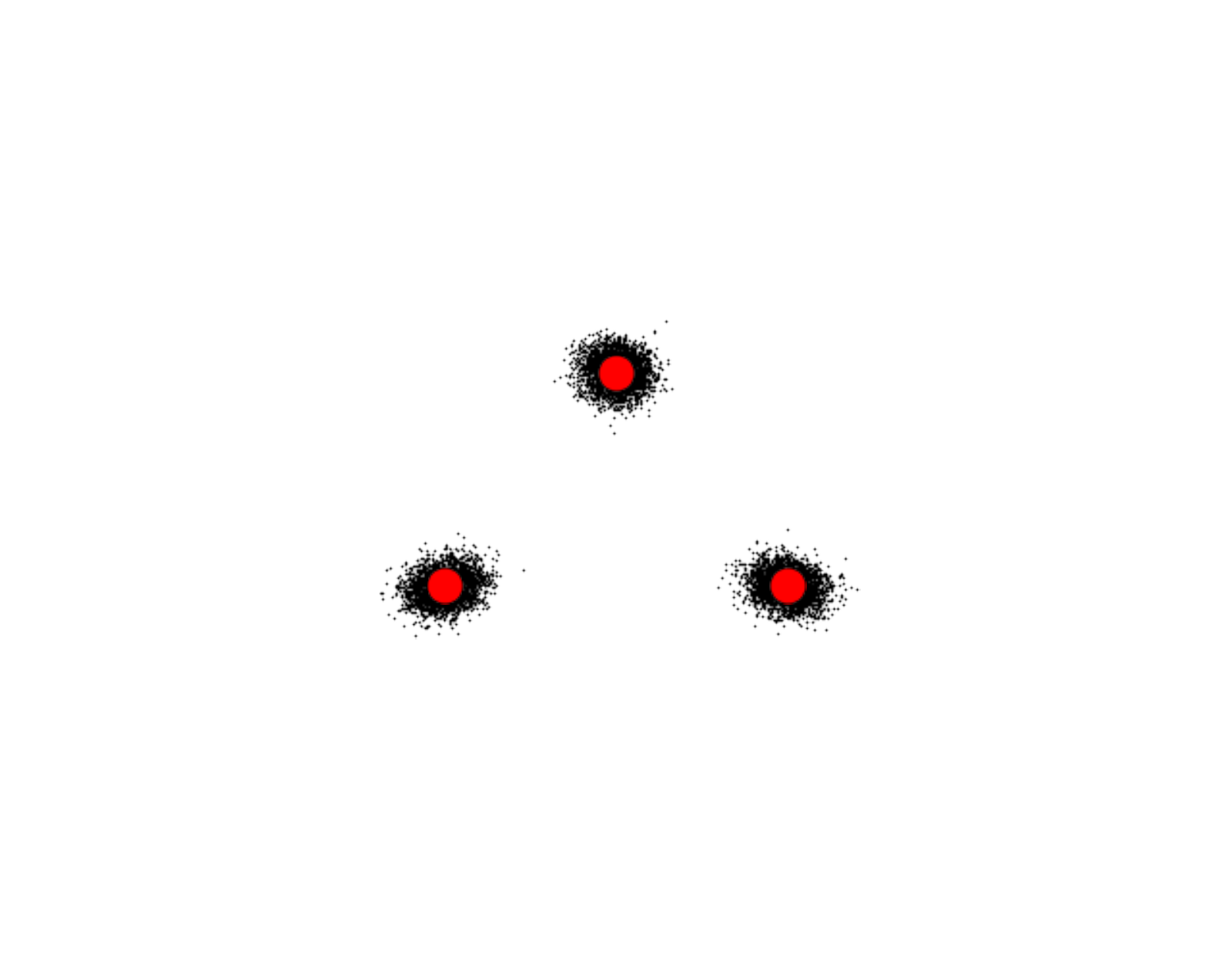}}
    \subfloat[SPX]{\includegraphics[width=\widthvar\textwidth]{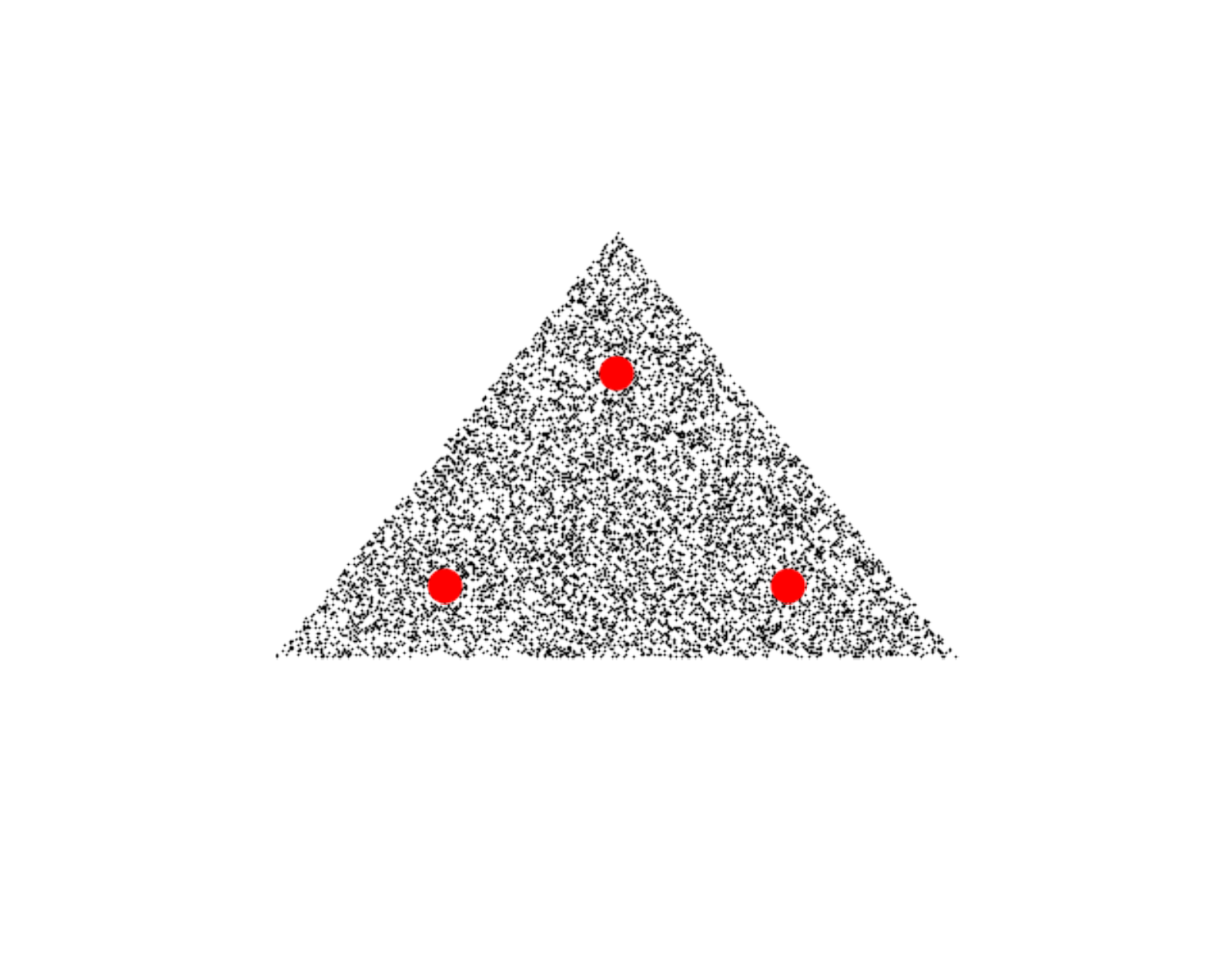}}
    \subfloat[REX]{\includegraphics[width=\widthvar\textwidth]{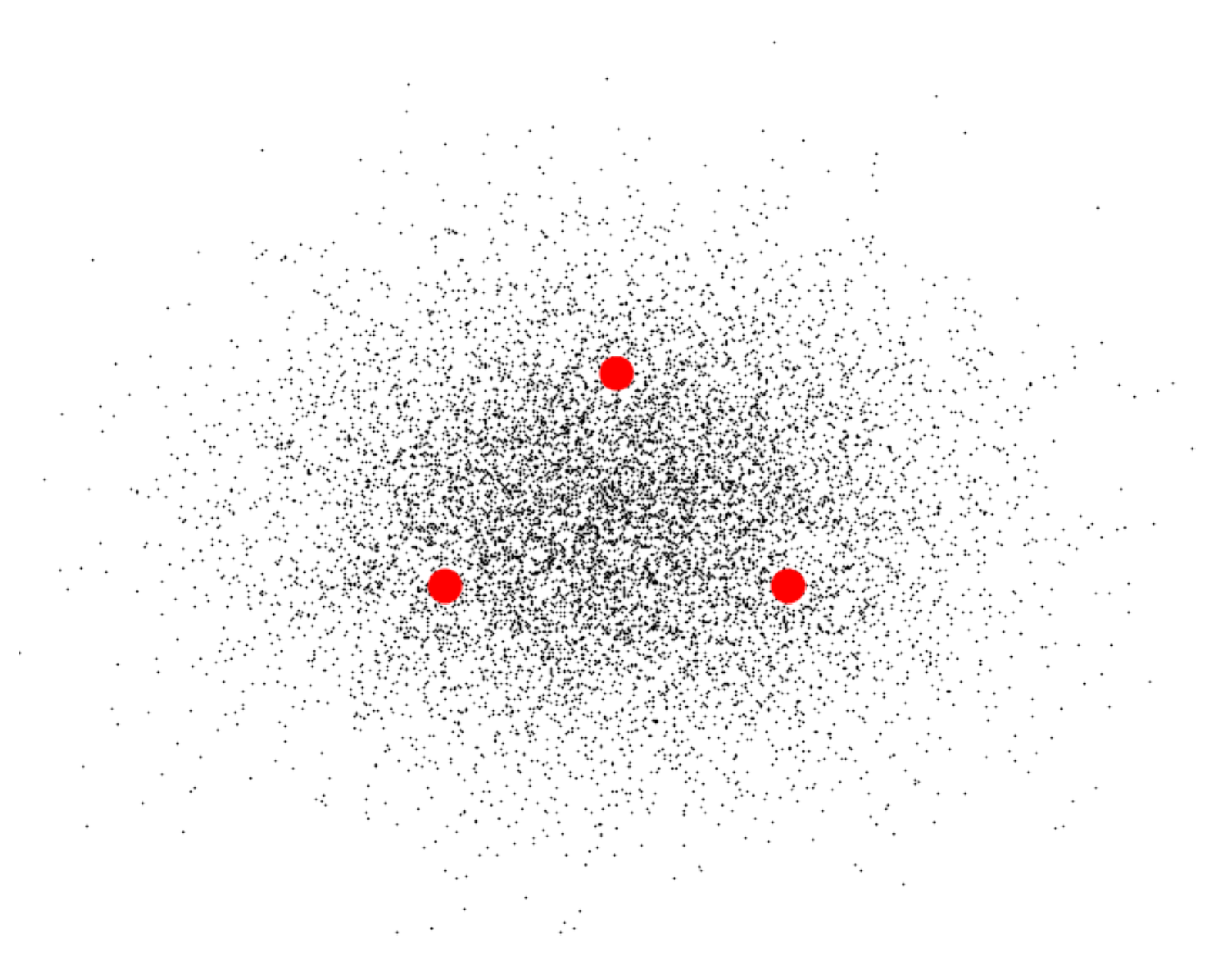}}
    \caption{
\small
Distribution of children generated by the five crossover methods.
Large red points are their parents.
%Distribution of children and parents the five crossover methods.
}
\label{fig:dist_children}
  \end{center}
\end{figure*}

%\label{fig:dist_children}

\section{Preliminaries}
\label{sec:preliminaries}

%% %\begin{defi}[MOPs]
\subsection{Definition of continuous MOPs}
\label{sec:def_MOPs}

% = (x_1, ..., x_D)^{\rm T}
%We del

A continuous MOP is to find a solution $\vector{x} \in \mathbb{S}$ that minimizes a given objective function vector $\vector{f}: \mathbb{R}^n \rightarrow \mathbb{R}^m, \vector{x} \mapsto \vector{f}(\vector{x})$.
Here, $\mathbb{S} \subseteq \mathbb{R}^n$ is the $n$-dimensional solution space, and $\mathbb{R}^m$ is the $m$-dimensional objective space.
$n$ is the number of decision variables, and $m$ is the number of objective functions.

%$\mathbb{S} = \prod^D_{j=1} [x^{\rm min}_j, x^{\rm max}_j]$ is the $n$-dimensional solution space where $x^{\rm min}_j \leq x_j \leq x^{\rm max}_j$ for each index $j \in \{1, ..., D\}$.
%Usually,  $\vector{f}$ consists of $M$ conflicting objective functions.
%Thus, no solution can simultaneously minimize all $M$ objective functions.

A solution $\vector{x}^{(1)}$ is said to dominate $\vector{x}^{(2)}$ iff $f_i (\vector{x}^{(1)}) \leq f_i (\vector{x}^{(2)})$ for all $i \in \{1, ..., m\}$ and $f_i (\vector{x}^{(1)}) < f_i (\vector{x}^{(2)})$ for at least one index $i$.
If $\vector{x}^*$ is not dominated by any other solutions in $\mathbb{S}$, $\vector{x}^*$ is a Pareto optimal solution.
The set of all $\vector{x}^*$ is the Pareto optimal solution set, and the set of all $\vector{f}(\vector{x}^*)$ is the Pareto front.
The goal of MOPs for the ``a posteriori'' decision making is to find a non-dominated solution set that approximates the Pareto front in the objective space.

%In general, the goal of MOPs for the ``a posteriori'' decision making is to find a set of non-dominated solutions that are well distributed and close to the Pareto front in the objective space.

%\cite{}

%% \subsection{Unbounded external archive}
%% \label{sec:uea}

\subsection{Crossover methods in real-coded GAs}

We use the following five crossover methods in real-coded GAs: simulated binary crossover (SBX) \cite{DebA95}, blend crossover (BLX) \cite{EshelmanS92}, parent-centric crossover (PCX) \cite{DebAJ02}, simplex crossover (SPX) \cite{TsutsuiYH99}, and real-coded ensemble crossover (REX) \cite{AkimotoSOK09}.
%Here, we explain the properties of the five crossover methods.
Here, we briefly explain the five crossover methods.

Traditional GAs use two variation operators: crossover and mutation.
In contrast, real-coded GAs with BLX, PCX, SPX, and REX do not need the mutation operator because they can generate diverse children by adjusting their control parameters (e.g., the expansion rate $\epsilon$ in SPX).
However, the polynomial mutation (PM) \cite{DebA95} is applied to two children generated by SBX in most studies.
In other words, SBX and PM have been considered to be a set.
For this reason, we apply PM to children generated only by SBX.
We refer to ``SBX and PM'' as ``SBX'' for simplicity.

%For this reason, we do not use any mutatation operator 

%% In the polynomial mutation , the $i$-th element of a child $\vector{u}$ ($\vector{u}_1$ or $\vector{u}_2$) is modified as follows:

%We forcus on the properties of the five crossover methods.

Table \ref{tab:cross_properties} shows the properties of the five crossover methods.
Although SBX and BLX are traditional two-parent crossover methods, PCX, SPX, and REX are multi-parent crossover methods.
PCX, SPX, and REX are rotationally invariant.
The performance of EMOAs with rotationally invariant operators does not depend on the coordinate system.
While PCX and REX use a Normal probability distribution, BLX and SPX use a uniform probability distribution.
The probability distribution used in SBX is unclear.
Although the center of the distribution of children is the mean vector of $k$ parents in BLX, SPX, and REX, that is one of the parents in SBX and PCX.
SPX and REX have a property called the ``preservation of statistics'' proposed in \cite{KitaOK98}.
In a crossover method with this property, children inherit the statistics (e.g., the mean vector and the covariance matrix) from their parents.

%random numbers

\begin{table}[t]
\renewcommand{\arraystretch}{1.0}
\begin{center}
  \caption{Properties of the five crossover methods, including the center of the distribution of children (parent or mean), the type of probability distribution (U: uniform or N: normal), the rotational invariance, the preservation of statistics, the number of parents $k$, and other control parameters.}
% Preservation of statistics
  %% {\scriptsize
          {\footnotesize
%{\small
  \label{tab:cross_properties}
\scalebox{0.98}[1]{  
\begin{tabular}{lccccccc}
\toprule
 & Cent. & Prob. & Rot. & Sta. & $k$ & Parameters \\
  \midrule
SBX & parent  & ? &  & & $2$ & $\eta_c = 20$, $\eta_m = 20$\\
BLX & mean & U & & & $2$ &  $\alpha = 0.5$\\
PCX & parent & N & $\checkmark$ &  & $3$ & $\sigma_{\zeta}^2 = 0.1$, $\sigma_{\eta}^2 = 0.1$\\
SPX & mean & U & $\checkmark$ & $\checkmark$ & $n+1$ & $\epsilon = \sqrt{n+2}$\\
REX & mean & N & $\checkmark$ & $\checkmark$ & $n+1$ & $\sigma^2 = 1/(k-1)$\\
%% SBX \cite{DebA95} & parent  & ? &  & & $2$ & $\eta_c = 20$, $\eta_m = 20$\\
%% BLX-$\alpha$ \cite{EshelmanS92} & mean & U & & & $2$ &  $\alpha = 0.5$\\
%% PCX \cite{DebAJ02} & parent & N & $\checkmark$ &  & $3$ & $\sigma_{\zeta}^2 = 0.1$, $\sigma_{\eta}^2 = 0.1$\\
%% SPX \cite{TsutsuiYH99} & mean & U & $\checkmark$ & $\checkmark$ & $n+1$ & $\epsilon = \sqrt{n+2}$\\
%% REX \cite{AkimotoSOK09} & mean & N & $\checkmark$ & $\checkmark$ & $n+1$ &  $\epsilon = 1/\sqrt{k-1}$\\
\toprule
\end{tabular}
}
}
\end{center}
\end{table}

Figure \ref{fig:dist_children} shows the distribution of children generated by the five crossover methods.
SBX simulates the working principle of the single-point crossover in binary-coded GAs.
Since SBX is a variable-wise operator, most children are generated along the coordinate axes.
The distribution of children is controlled by $\eta_c$ in SBX (and $\eta_m$ in PM).
In BLX, the $j$-th element ($j \in \{1, ..., n\}$) of a child is uniformly randomly selected from the range $[l_j, u_j]$.
Here, $l_j = \min(x^{(1)}_{j}, x^{(2)}_{j}) - \alpha |x^{(1)}_{j} - x^{(2)}_{j}|$ and $u_j = \max(x^{(1)}_{j}, x^{(2)}_{j}) + \alpha |x^{(1)}_{j} - x^{(2)}_{j}|$.
$\vector{x}^{(1)}$ and $\vector{x}^{(2)}$ are parents, and $\alpha$ is the expansion factor.

%% UNDX-$m$ \cite{KitaOK99} is a multi-parent extension of unimodal normal distributed crossover (UNDX) \cite{OnoK97}.
%% PCX is a parent-centric version of UNDX-$m$.

PCX is a parent-centric version of UNDX-$m$ \cite{KitaOK99}, which is a multi-parent extension of unimodal normal distribution crossover (UNDX) \cite{OnoK97}.
While the center of the distribution of children is the mean vector of parents in UNDX-$m$, that is one of the parents in PCX.
PCX requires two parameters $\sigma_{\zeta}^2$ and $\sigma_{\eta}^2$ that control the variances of two Normal distributions.
SPX can be viewed as being a rotationally invariant version of BLX.
SPX uniformly generates children inside an expanded simplex formed by $k$ parents. % ($k=n+1$).
The theoretical analysis presented in \cite{HiguchiTY00} shows that SPX with the expansion factor $\epsilon = \sqrt{n+2}$ satisfies the preservation of statistics.
REX is a generalized version of UNDX-$m$.
REX using a zero-mean Normal distribution with the variance $\sigma^2 = 1/(k-1)$ satisfies the preservation of statistics \cite{Akimoto10}.

\subsection{Environmental selections}
\label{sec:env_selection}

%"indices are mutually different"
% = \{\vector{x}^{(1)}, ..., \vector{x}^{(\mu)}\}

We consider a ``simple'' EMOA shown in Algorithm \ref{alg:emoa}.
After the initialization of the population $\vector{P}$ with the population size $\mu$ (line 1), the following operations are repeatedly performed until a termination condition is satisfied.
First, $k$ parents are randomly selected from $\vector{P}$ such that their indices are different from each other (line 3).
Let $\vector{R}$ be a set of the $k$ parents.
Then, $\lambda$ children are generated by applying a crossover method to the same $k$ parents $\lambda$ times (line 4).\footnote{Since SBX generates two children in a single operation, SBX is performed $\lambda/2$ times.}
To effectively exploit the neighborhood of the $k$ parents, the same parents are generally used to generate children in GAs for single-objective optimization \cite{Akimoto10}.
Let $\vector{Q}$ be a set of the $\lambda$ children.
At the end of each iteration, the environmental selection is performed using $\vector{P}$, $\vector{R}$, and $\vector{Q}$ (line 5).

Below, we explain the following three environmental selections: best-all (BA), best-family (BF), and best-children (BC).
Note that our main contributions in this paper are analysis of BA, BF, and BC in Section \ref{sec:experimental_results}, not proposing BA, BF, and BC.
Algorithms \ref{alg:ba}, \ref{alg:bf}, and \ref{alg:bc} show BA, BF, and BC, respectively.
While BA and BF are elitist selections, BC is a non-elitist selection.
The three selections require a method of ranking individuals based on their quality.
Similar to MO-CMA-ES \cite{IgelHR07}, BA, BF, and BC can be combined with any ranking method.
In this paper, we use four ranking methods in NSGA-II \cite{DebAPM02}, SMS-EMOA \cite{BeumeNE07}, SPEA2 \cite{ZitzlerLT01}, and IBEA with the additive $\epsilon$ indicator \cite{ZitzlerK04}.
We denote their ranking methods as ``NS'', ``SM'', ``SP'', and ``IB'', respectively.
Individuals are ranked based on their non-domination levels in NS and SM.
The tie-breakers are the crowding distance in NS and the hypervolume contribution in SM.
In SP and IB, individuals are sorted based on their so-called fitness values in descending order.
In this paper, X-Y-Z represents the EMOA (Algorithm \ref{alg:emoa}) with an environmental selection X, a crossover method Y, and a ranking method Z.
For example, BA-SBX-NS is the EMOA with BA, SBX, and NS.

%,VossHI10

%Ranks of individuals are determined based on their so-called fitness values in the ranking methods in SP and IB.

%% Individuals are ranked based on their non-domination levels and crowding distance values in the ranking method in NS.
%% Also, individuals are ranked based on their non-domination levels and hypervolume contributions in the ranking method in SM.
%The so-called fitness values are used to assign ranks to individuals in the ranking methods in SPEA2 \cite{ZitzlerLT01} and IBEA \cite{ZitzlerK04}.

\IncMargin{0.5em}
\begin{algorithm}[t]
%\scriptsize
%\footnotesize
\small
%\SetAlgoLined
\SetSideCommentRight
%\KwData{this text}
%\KwResult{how to write algorithm with \LaTeX2e }
$t \leftarrow 1$, initialize the population $\vector{P} =\{ \vector{x}^{(1)}, ..., \vector{x}^{(\mu)}\}$\;
\While{$\textsf{\upshape{The termination criteria are not met}}$}{
  $\vector{R} \leftarrow$ Randomly select $k$ parents from $\vector{P}$\;
  $\vector{Q} \leftarrow$ Generate $\lambda$ children by applying the crossover method to $\vector{R}$\;
  $\vector{P} \leftarrow {\rm environmentalSelection(\vector{P}, \vector{Q}, \vector{R})}$\;
  $t \leftarrow t + 1$\;
}
%
%% \KwRet $\vector{P}$ for the decision making (Subsection \ref{sec:decision_making}) or benchmarking (Subsection \ref{sec:ada_postprocessing})\;
\caption{The simple EMOA}
\label{alg:emoa}
\end{algorithm}\DecMargin{0.5em}

\IncMargin{0.5em}
\begin{algorithm}[t]
%\scriptsize
%\footnotesize
\small
%\SetAlgoLined
\SetSideCommentRight
%\KwData{this text}
%\KwResult{how to write algorithm with \LaTeX2e }
%$\vector{P} \leftarrow \vector{P} \backslash \vector{R}$\;
Assign ranks to all individuals in $\vector{P} \cup \vector{Q}$\;
$\vector{S} \leftarrow \vector{P} \cup \vector{Q}$ and $\vector{P} \leftarrow \emptyset$\;
\For{$i \in \{1, ..., \mu\}$}{
  $\vector{x} \leftarrow$ Select the best ranked individual from $\vector{S}$\;
  $\vector{P} \leftarrow \vector{P} \cup \{\vector{x}\}$ and  $\vector{S} \leftarrow \vector{S} \setminus \{\vector{x}\}$\;
}
%
%% \KwRet $\vector{P}$ for the decision making (Subsection \ref{sec:decision_making}) or benchmarking (Subsection \ref{sec:ada_postprocessing})\;
\caption{BA (the elitist selection)}
\label{alg:ba}
\end{algorithm}\DecMargin{0.5em}

\IncMargin{0.5em}
\begin{algorithm}[t]
%\scriptsize
%\footnotesize
\small
%\SetAlgoLined
\SetSideCommentRight
%\KwData{this text}
%\KwResult{how to write algorithm with \LaTeX2e }
%$\vector{S} \leftarrow \vector{P} \cup \vector{Q}$\;
Assign ranks to all individuals in $\vector{P} \cup \vector{Q}$\;
$\vector{S} \leftarrow \vector{Q} \cup \vector{R}$ and $\vector{P} \leftarrow \vector{P} \setminus \vector{R}$\;
\For{$i \in \{1, ..., k\}$}{
  $\vector{x} \leftarrow$ Select the best ranked individual from $\vector{S}$\;
  $\vector{P} \leftarrow \vector{P} \cup \{\vector{x}\}$ and $\vector{S} \leftarrow \vector{S} \setminus \{\vector{x}\}$\;
}
%
%% \KwRet $\vector{P}$ for the decision making (Subsection \ref{sec:decision_making}) or benchmarking (Subsection \ref{sec:ada_postprocessing})\;
\caption{BF (the elitist restricted selection)}
\label{alg:bf}
\end{algorithm}\DecMargin{0.5em}

\IncMargin{0.5em}
\begin{algorithm}[t]
%\scriptsize
%\footnotesize
\small
%\SetAlgoLined
\SetSideCommentRight
%\KwData{this text}
%\KwResult{how to write algorithm with \LaTeX2e }
$\vector{P} \leftarrow \vector{P} \setminus \vector{R}$\;
Assign ranks to all individuals in $\vector{P} \cup \vector{Q}$\;
%$\vector{S} \leftarrow \vector{Q} \cup \vector{R}$ and $\vector{P} \leftarrow \vector{P} \setminus \vector{R}$\;
\For{$i \in \{1, ..., k\}$}{
  $\vector{x} \leftarrow$ Select the best ranked individual from $\vector{Q}$\;
  $\vector{P} \leftarrow \vector{P} \cup \{\vector{x}\}$ and $\vector{Q} \leftarrow \vector{Q} \setminus \{\vector{x}\}$\;
}

%
%% \KwRet $\vector{P}$ for the decision making (Subsection \ref{sec:decision_making}) or benchmarking (Subsection \ref{sec:ada_postprocessing})\;
\caption{BC (the non-elitist restricted selection)}
\label{alg:bc}
\end{algorithm}\DecMargin{0.5em}

%% In BA, the best $N$ individuals regarding the ranks of individuals are selected from the union of $\vector{P}$ and $\vector{Q}$.
%% the best among all individauls (BA), the best among a family (BF), and the best among children (BC).

In BA (Algorithm \ref{alg:ba}), the top-ranked $\mu$ individuals are selected from the union of $\vector{P}$ and $\vector{Q}$.
BA is the traditional elitist $(\mu+\lambda)$-selection used in most EMOAs (e.g., NSGA-II and SPEA2).
It should be noted that BA-SBX-NS is not identical to NSGA-II.
The differences between BA-SBX-NS and NSGA-II are the random parent selection and the children generation.
The same $k$ parents are used to generate $\lambda$ children in BA.
For the same reason, BA-SBX-SP, BA-SBX-SM, and BA-SBX-IB are not identical to SPEA2, SMS-EMOA, and IBEA, respectively.

%The same is true for other EMOAs (SPEA2, SMS-EMOA, and IBEA).

%% In Algorithm \ref{alg:emoa}, the same 
%% random parent selection method
%BA can be viewed as the naive extension of existing EMOAs to the multiple offspring sampling framework.

In BF (Algorithm \ref{alg:bf}), the environmental selection is performed only among the so-called ``family'' that consists of $\lambda$ children in $\vector{Q}$ and $k$ parents in $\vector{R}$.
After all individuals in the union of $\vector{P}$ and $\vector{Q}$ have been ranked, only $k$ parents in $\vector{R}$ are removed from $\vector{P}$.
Then, the best $k$ individuals are selected from the union of $\vector{Q}$ and $\vector{R}$.
Although non-parents in $\vector{P}$ do not directly participate in the selection process, they contribute to assign ranks to individuals in the union of $\vector{Q}$ and $\vector{R}$.
While the maximum number of individuals replaced by children is $\mu$ in BA, that is $k$ in BF.
Since only $k$ parents can be replaced by children in BF, non-parents can survive to the next iteration with no comparison.
Selections among families as in BF are used in GAs for single-objective optimization (e.g., the deterministic crowding \cite{Mahfoud92}).

%Environmental selections among families similar to BF are used in GAs for single-objective optimization (e.g., the deterministic crowding \cite{Mahfoud92} and the minimal generation gap \cite{SatohYK96}).

%Non-parents can survive to the next iteration.
%By restricting the replacement of individuals, BF prohibits that other non-parents are replaced by children.
%MGG
%Mahfoud
%% BF prevents that inferior non-parent individauls in $\vector{P}$ are deleted by the $\lambda$ children.
%% Thus, the convergence ability towards the Pareto front of BF should be weaker than BA.

In BC (Algorithm \ref{alg:bc}), the environmental selection is performed among $\lambda$ children in $\vector{Q}$.
We assume that $\lambda \geq k$.
After $k$ parents in $\vector{R}$ have been removed from $\vector{P}$, all individuals in the union of $\vector{P}$ and $\vector{Q}$ are ranked.
Then, the best $k$ individuals are selected from $\vector{Q}$.
Since all $k$ parents are deleted from $\vector{P}$ regardless of their quality, BC does not maintain non-dominated individuals in $\vector{P}$.
Thus, BC is a non-elitist selection in contrast to the elitist BA and BF selections.
While $\mu$ individuals in $\vector{P}$ are replaced with $\lambda$ children in $\vector{Q}$ in most classical $(\mu, \lambda)$-EMOAs (e.g., MOGA), only $k$ parents in $\vector{R}$ are replaced with the best $k$ out of $\lambda$ children in $\vector{Q}$ in BC.
Thus, BC is different from the traditional $(\mu, \lambda)$-selection.

BC can be viewed as being an extension of just generation gap (JGG) \cite{Akimoto10} to multi-objective optimization.
JGG is an environmental selection in GAs for single-objective continuous optimization.
The only difference between BC and JGG is how to assign ranks to individuals.
Individuals $\vector{x}^{(1)}, \vector{x}^{(2)}, ...$ are ranked based on their objective values $f(\vector{x}^{(1)}), f(\vector{x}^{(2)}), ...$ in JGG and their objective vectors $\vector{f}(\vector{x}^{(1)}), \vector{f}(\vector{x}^{(2)}), ...$ in BC.
The results presented in \cite{Akimoto10} show that non-elitist GAs with JGG significantly outperform elitist GAs on single-objective test problems (especially multimodal problems) when using crossover methods with the preservation of statistics.

%% file: experimental_settings.tex
\section{Experimental settings}
\label{sec:experimental_settings}

%This section describes experimental settings.
%
We conducted all experiments using the comparing continuous optimizers (COCO) platform \cite{HansenAMTB16}.
COCO is the standard platform used in the black box optimization benchmarking (BBOB) workshops held at GECCO (2009--present).
We used the latest COCO software (version 2.2.2) downloaded from \url{https://github.com/numbbo/coco}.
COCO provides six types of BBOB problem suites, including the single-objective BBOB noiseless problem suite \cite{HansenFRA09}.
%The single-objective BBOB noiseless problem suite \cite{HansenFRA09} consists of well-understood 24 test problems which are grouped into 5 categories: (i) separable functions ($f^{\rm single}_1, ..., f^{\rm single}_{5}$), (ii) functions with low or moderate conditioning ($f^{\rm single}_6, ..., f^{\rm single}_{9}$), (iii) functions with high conditioning and unimodal ($f^{\rm single}_{10}, ..., f^{\rm single}_{14}$), (iv) multimodal functions with adequate global structure ($f^{\rm single}_{15}, ..., f^{\rm single}_{19}$), and (v) multimodal functions with weak global structure ($f^{\rm single}_{20}, ..., f^{\rm single}_{24}$).
%
The bi-objective BBOB problem suite \cite{TusarBHA16} consists of 55 bi-objective test problems $\vector{f}_1, ..., \vector{f}_{55}$ designed based on the idea presented in \cite{BrockhoffTH15}.
Each bi-objective BBOB problem is constructed by combining two single-objective BBOB problems.
For example, the first and second objective functions of $\vector{f}_7$ are the Sphere function and the rotated Rastrigin function, respectively.
The number of decision variables $n$ is $n \in \{2, 3, 5, 10, 20, 40\}$.
For details of the 55 bi-objective test problems, see \cite{TusarBHA16}.
For each problem, $15$ runs were performed.
These settings adhere to the analysis procedure adopted by the GECCO BBOB community.
The maximum number of function evaluations was set to $10^4 \times n$.

COCO also provides the post-processing tool that aggregates experimental data.
COCO automatically stores all non-dominated solutions found by an optimizer in the unbounded external archive.
The performance indicator $I_{\rm COCO}$ \cite{BrockhoffTTWHA16} in COCO  is mainly based on the hypervolume value of non-dominated solutions in the unbounded external archive.
When no solution in the external archive dominates a predefined reference point in the normalized objective space, the $I_{\rm COCO}$ value is calculated based on the distance to the so-called region of interest.
For details of $I_{\rm COCO}$, see \cite{BrockhoffTTWHA16}.

%\cite{ZitzlerT98}
%% s, the code provides a platform to benchmark and compare continuous optimizers,

%% We use the bi-objective BBOB problem suite .
%% The BBOB suite consists of 55 various bi-objective test problems $f_1, ..., f_{55}$.
%% The 55 functions were carefully designed based on the idea presented in \cite{BrockhoffTH15}.
%% The each function consists of two single-objective BBOB  functions \cite{}.

%\footnote{Source codes of all algorithms and experimental data will be uploaded to the author's website if this paper is accepted for publication.}
%We examine the performance of an EMOA (Algorithm \ref{alg:emoa}) with the three environmental selections (BA, BF, and BC).
We implemented all algorithms using jMetal \cite{DurilloN11}.
Source codes of all algorithms are available at \url{https://sites.google.com/view/nemorgecco2019/}.
For all five crossover methods (except for PCX), we used the control parameters recommended in the literature shown in Table \ref{tab:cross_properties}.
Since PCX with $k=3$ performed poorly in our preliminary study, we set $k$ to $n+1$ similar to SPX and REX.
For comparison, we evaluated the performance of the original NSGA-II, SPEA2, SMS-EMOA, and IBEA.
SBX and PM with $p_c = 0.9$, $\eta_c = 20$, $p_m = 1/n$, and $\eta_m = 20$ were used in the original EMOAs.
As in \cite{TusarF16}, $\mu$ was set to $\lfloor 100\,{\rm ln}(n)\rfloor$.
The number of children $\lambda$ was set to $10n$.
We set the $\lambda$ value based on our preliminary results and studies of GAs for single-objective optimization (e.g., \cite{AkimotoSOK09,Akimoto10}).

%the population size

%No restarts are performed.

%In our preliminary results, EMOAs with $\mu=100\lfloor{\rm ln}(n)\rfloor$ achieve better results than those with $\mu=100$ on the bi-objective BBOB problem suite.
%We set the maximum number of function evaluations to $10\,000 \times n$.

% \subsubsection{Test problems.}
%% \subsubsection{Performance indicators.}

%% file: experimental_results.tex
\begin{figure*}[t]
\newcommand{\widthvar}{0.325}
  \begin{center} 
\subfloat[$n=2$]{\includegraphics[width=\widthvar\textwidth]{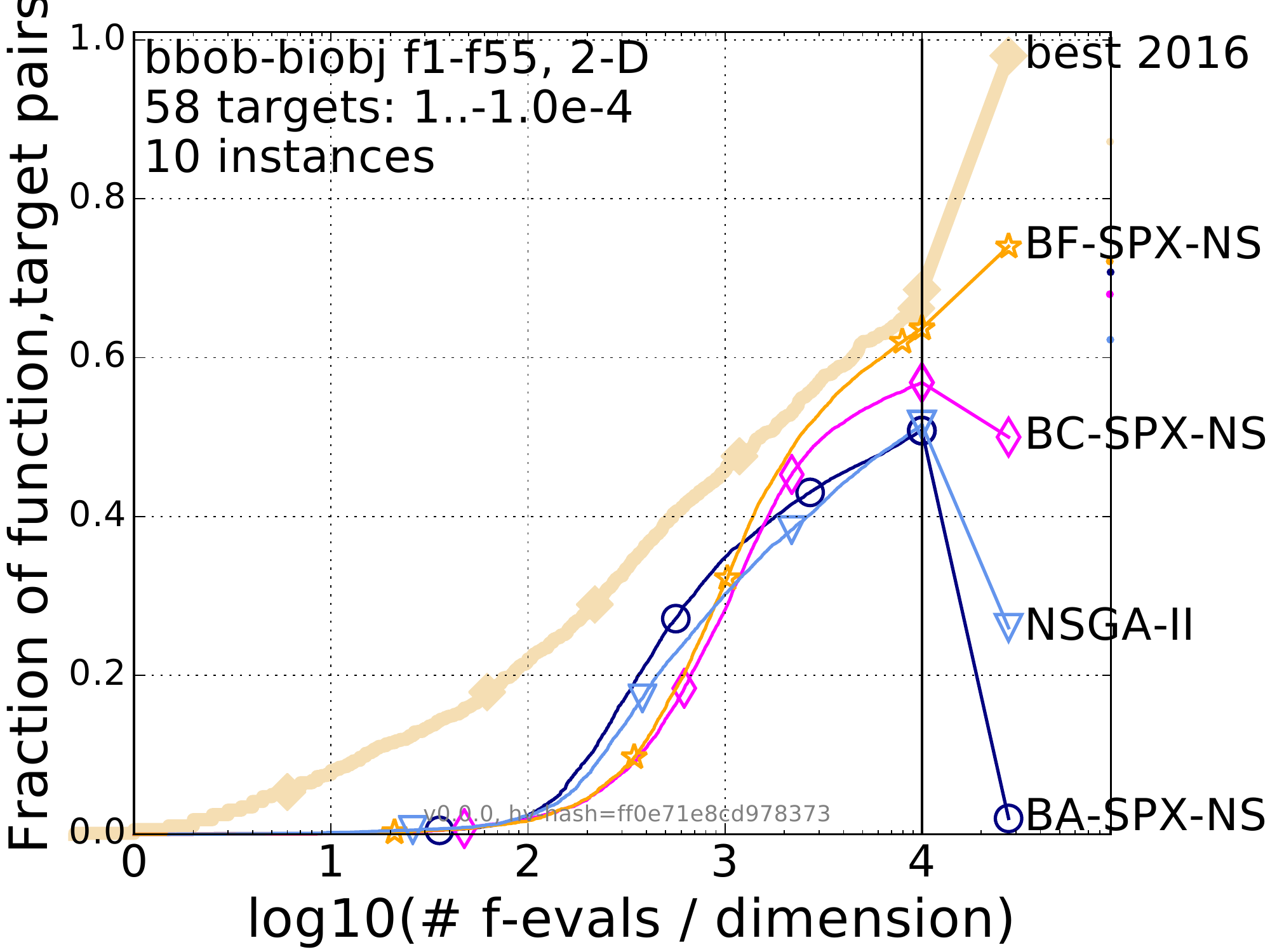}}
%% \subfloat[$n=3$]{\includegraphics[width=\widthvar\textwidth]{graph/spx_nsgaii/pprldmany_03D_noiselessall.pdf}}
%% \subfloat[$n=5$]{\includegraphics[width=\widthvar\textwidth]{graph/spx_nsgaii/pprldmany_05D_noiselessall.pdf}}
%% \\
\subfloat[$n=10$]{\includegraphics[width=\widthvar\textwidth]{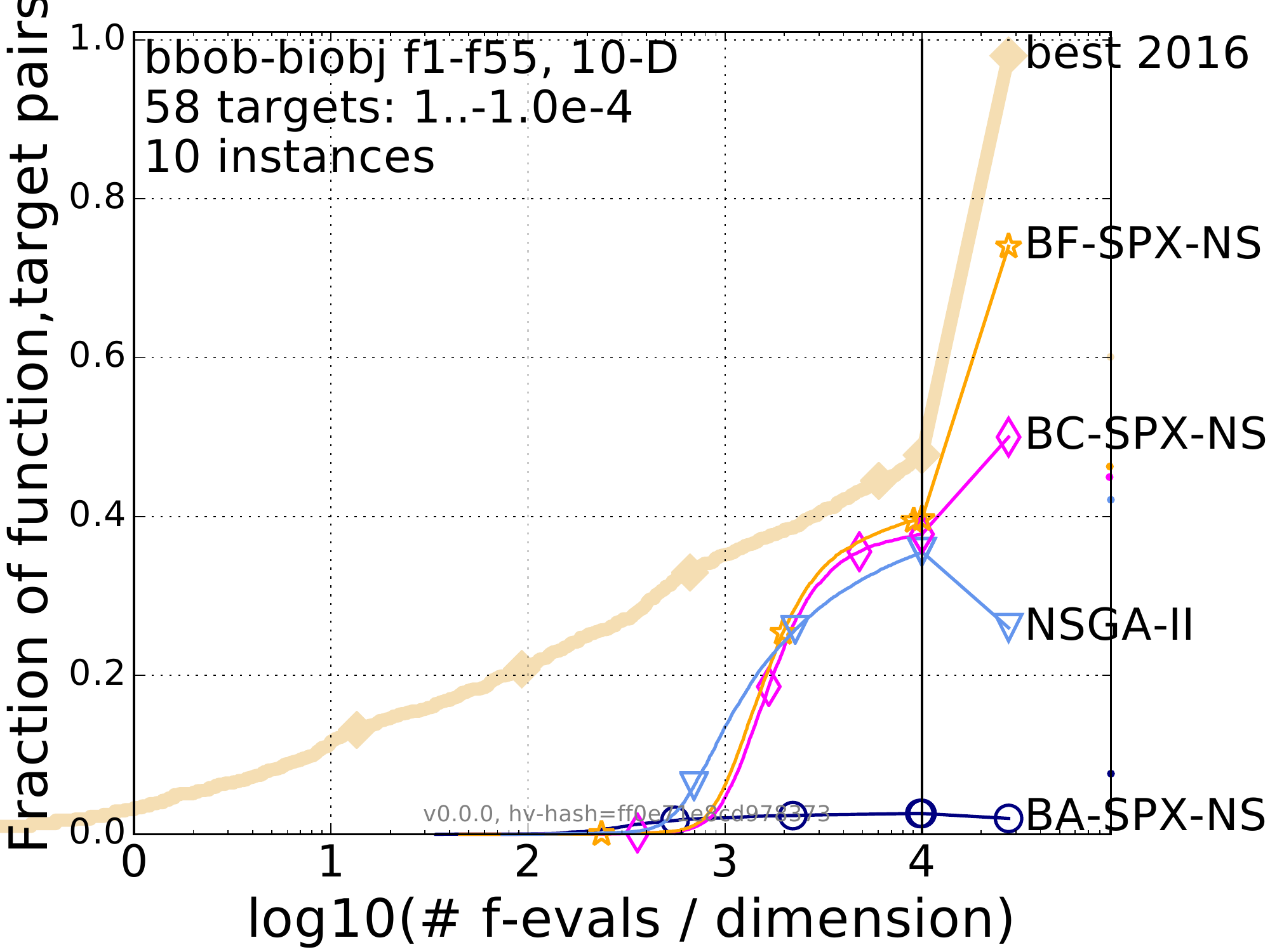}}
%\subfloat[$n=20$]{\includegraphics[width=\widthvar\textwidth]{graph/spx_nsgaii/pprldmany_20D_noiselessall.pdf}}
\subfloat[$n=40$]{\includegraphics[width=\widthvar\textwidth]{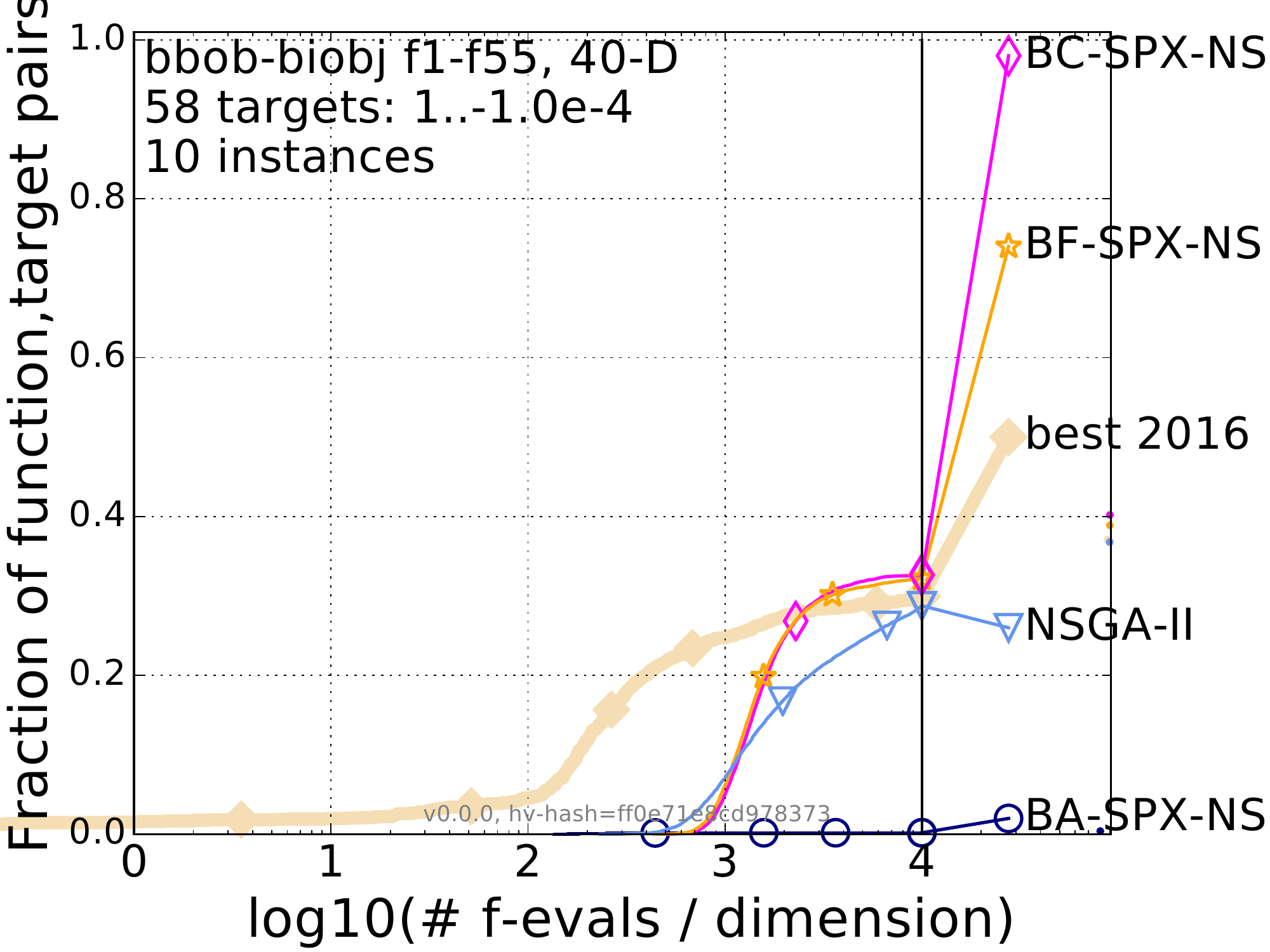}}
    \caption{
\small
Results of the original NSGA-II, BA-SPX-NS, BF-SPX-NS, and BC-SPX-NS on all 55 bi-objective BBOB test problems with $n \in \{2, 10, 40\}$ (higher is better).
For the notation X-Y-Z, see Subsection \ref{sec:env_selection}.
    }
\label{fig:emoa_spx_nsgaii}
  \end{center}
\end{figure*}

\section{Results}
\label{sec:experimental_results}

%% Results from experiments according to \cite{hansen2016exp},
%% \cite{hansen2016perfass} and \cite{biobj2016perfass} on the benchmark
%% functions given in \cite{biobj2016func} are presented in
%% Figures~\ref{fig:ECDFsingleOne}, \ref{fig:ECDFsingleTwo}, \ref{fig:ECDFsGroupsFive} and
%% \ref{fig:ECDFsGroupsTwenty} and in Tables~\ref{tab:aRTs5} and~\ref{tab:aRTs20}.
%% The experiments were performed with COCO \cite{hansen2016cocoplat}, version
%% \change{2.0}, the plots were produced with version \change{2.0}.

%% The \textbf{average runtime (aRT)}, used in the %figures and
%% tables,
%% depends on a given quality indicator value, $\Itarget=\hvref+\DI$, and is
%% computed over all relevant trials as the number of function
%% evaluations executed during each trial while the best indicator value
%% did not reach \Itarget, summed over all trials and divided by the
%% number of trials that actually reached \Itarget\
%% \cite{hansen2016exp,price1997dev}.  \textbf{Statistical significance}
%% is tested with the rank-sum test for a given target $\Itarget$
%% using, for each trial,
%% either the number of needed function evaluations to reach
%% $\Itarget$ (inverted and multiplied by $-1$), or, if the target
%% was not reached, the best $\DI$-value achieved, measured only up to
%% the smallest number of overall function evaluations for any
%% unsuccessful trial under consideration.

This section shows analysis of the three environmental selections (BA, BF, and BC).
Since SPX is suitable for BF and BC, we mainly discuss results of EMOAs with SPX.
Although results of EMOAs with REX are similar to those with SPX, we do not show them here due to space constraint.
As shown in Subsection \ref{sec:other_crossovers}, SBX, BLX, and PCX are not suitable for BA, BF, and BC.
%Also, any crossover method is not suitable for BA.

Subsection \ref{sec:comparison_three_selections} shows a comparison among BA-SPX-NS, BF-SPX-NS, BC-SPX-NS, and the original NSGA-II.
Subsection \ref{sec:analysis_ba} investigates why BA performs poorly.
Subsection \ref{sec:advantage_bc} analyzes the advantages and disadvantages of the non-elitist BC compared with the elitist BF.
Subsection \ref{sec:other_crossovers} examines the performance of BA, BF, and BC with other crossover methods (SBX, BLX, PCX, and REX).
Subsection \ref{sec:other_enviromental_selections} presents a comparison of BA, BF, and BC with other ranking methods (SP, SM, and IB).

\subsection{Comparison of BA, BF, and BC}
\label{sec:comparison_three_selections}

%% Analyssi: why BA does not work well?
%$\{-10^{-4}, -10^{-4.2}, -10^{-4.4}, -10^{-4.6}, -10^{-4.8}, -10^{-5}, 0, 10^{-5}, 10^{-4.9}, 10^{-4.8}, ..., 10^{-0.1}, 10^{0}$
%3, 5, 

Figure \ref{fig:emoa_spx_nsgaii} shows results of the original NSGA-II, BA-SPX-NS, BF-SPX-NS, and BC-SPX-NS on all 55 BBOB problems with $n \in \{2, 10, 40\}$.
Due to space constraint, results for $n \in \{3, 5, 20\}$ are not shown, but they are similar to results for $n \in \{2, 10\}$.
In this section, we use the SPX crossover and the NS ranking method.
In Figure \ref{fig:emoa_spx_nsgaii}, ``best 2016'' is a {\em virtual algorithm portfolio} that is constructed from the performance data of 15 algorithms participating in the GECCO BBOB 2016 workshop.
Note that ``best 2016'' does not mean the best optimizer among the 15 algorithms.

Figure \ref{fig:emoa_spx_nsgaii} shows the bootstrapped empirical cumulative distribution (ECDF) of the number of function evaluations (FEvals) divided by $n$ (FEvals/$n$) for 58 target $I_{\rm COCO}$ indicator values $\{-10^{-4},$ $ -10^{-4.2},$ $..., 10^{-0.1}, 10^{0}\}$ for all 55 BBOB problems with each $n$.
We used the COCO software to generate all ECDF figures in this paper.
In Figure \ref{fig:emoa_spx_nsgaii}, the vertical axis indicates the proportion of target $I_{\rm COCO}$ indicator values which a given optimizer can reach within specified function evaluations. 
For example, in Figure \ref{fig:emoa_spx_nsgaii} (b), BF-SPX-NS reaches about 40 percent of all 58 target $I_{\rm COCO}$ indicator values within $10^4 \times n$ evaluations on all 55 problems with $n=10$ in all runs.
If an optimizer finds all Pareto optimal solutions on all 55 problems in all runs, the vertical value becomes 1.
More detailed explanations of the ECDF (including illustrative examples) are found in \cite{BrockhoffTH15,BrockhoffTTWHA16}.

\textbf{Statistical significance} is also tested with the rank-sum test ($p=0.05$) for a given target value using the COCO software.
However, statistical test results are almost consistent with ECDF figures.
Additionally, the space of this paper is limited.
For these reasons, we show only ECDF figures.
The statistical test results and other ECDF figures are available at \url{https://sites.google.com/view/nemorgecco2019/}.

%% As the standard results for benchmarking, Figure \ref{fig:emoa_spx_nsgaii} also shows results of the original NSGA-II, whose components and parameter settings are exactly the same with the original experiment in \cite{DebAPM02} (except for $N$),
%% In our preliminary results, NSGA-II with $N=100\lfloor{\rm ln}(d)\rfloor$ achieve better results than that with $N=100$.

%According to the suggestion in \cite{BrockhoffTH15}, we mainly discuss results based on the anytime performance of EMOAs, rather than the final results.

Figure \ref{fig:emoa_spx_nsgaii} shows that BA-SPX-NS performs the best until $10^3 \times n$ evaluations for $n=2$.
However, the increase of $n$ deteriorates the performance of BA-SPX-NS.
The evolution of BA-SPX-NS clearly stagnates for $n \geq 10$.
The original NSGA-II is the best performer in the early stage for $n\geq 10$.
BF-SPX-NS and BC-SPX-NS perform better than NSGA-II and BA-SPX-NS in the later stage for all $n$.
Interestingly, the non-elitist BC-SPX-NS performs the best in the later stage for $n = 40$.
Although it has been believed that elitist EMOAs always outperform non-elitist EMOAs for about two decades, our results show that the non-elitist BC-SPX-NS performs better than the elitist NSGA-II, BA-SPX-NS, and BF-SPX-NS on the bi-objective BBOB problems with $n = 40$ when using the unbounded external archive.

%Although BF-SPX-NS outperforms BC-SPX-NS for $n \leq 20$, BC-SPX-NS performs better than BF-SPX-NS for $n = 40$.
%As far as we know,
%This result indicates that 

%These results also indicate that the restricted selections (BC and BF) and SPX can improve the performance of the original NSGA-II.
%That is, the non-elitist EMOAupdate method performs better than the elitist one.

%As $n$ increases, the difference between the performance of BF-SPX-NS and BC-SPX-NS get smaller.
%in the later search stage for all $n$.
%Figure \ref{fig:spx_nsgaii_each} shows results of NSGA-II, BA-SPX-NS, BF-SPX-NS, and BC-SPX-NS on $f_{54}$ and $f_{55}$ with $n=40$.
%We do not claim that BC always outperforms BF for all test problems with $n=40$.

\begin{figure}[t]
\newcommand{\widthvar}{0.23}
  \begin{center} 
    \subfloat[$\vector{f}_{54}$: BF outperforms BC]{
      \includegraphics[width=\widthvar\textwidth]{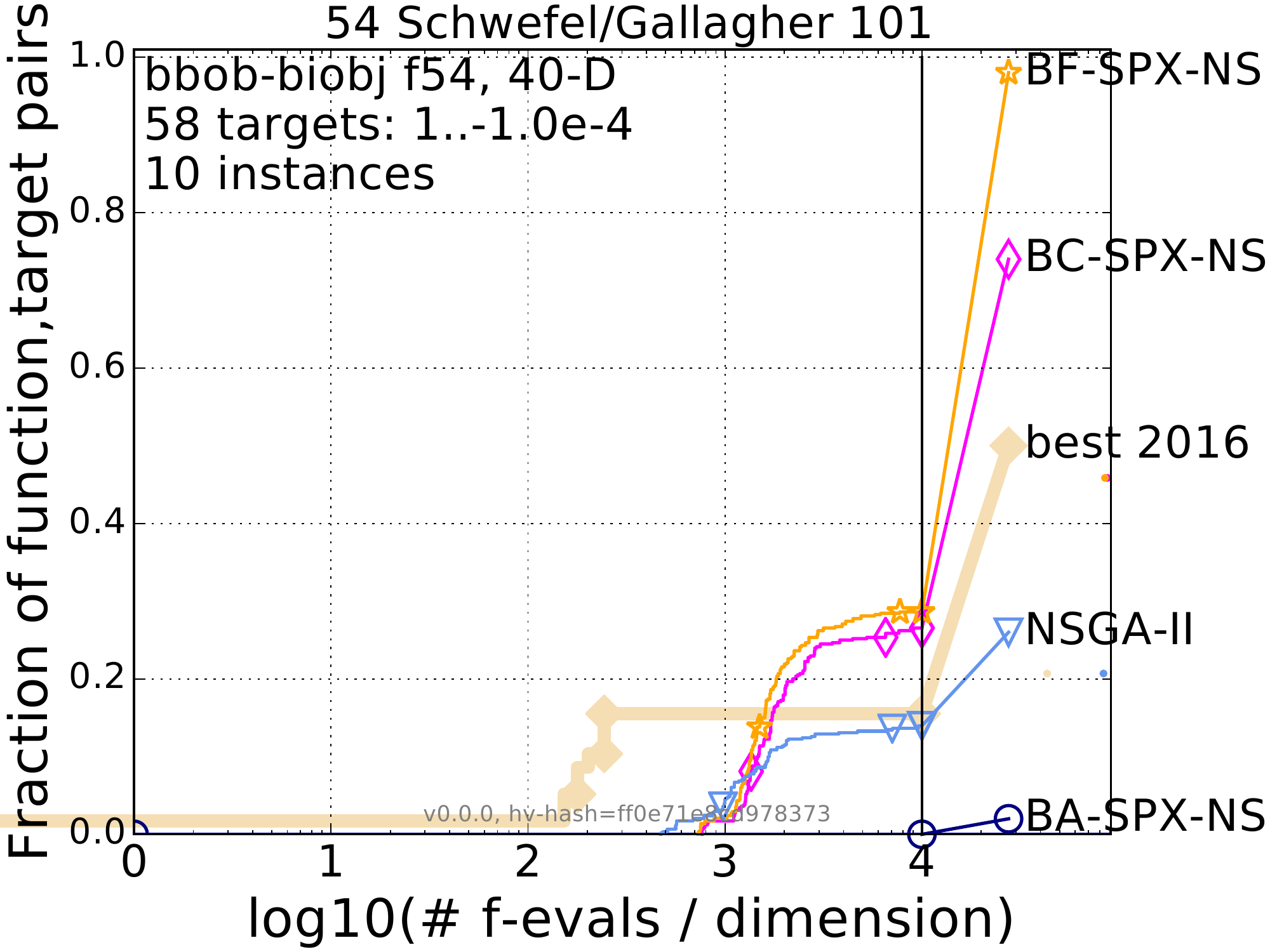}
    }
    \subfloat[$\vector{f}_{55}$: BC outperforms BF]{
      \includegraphics[width=\widthvar\textwidth]{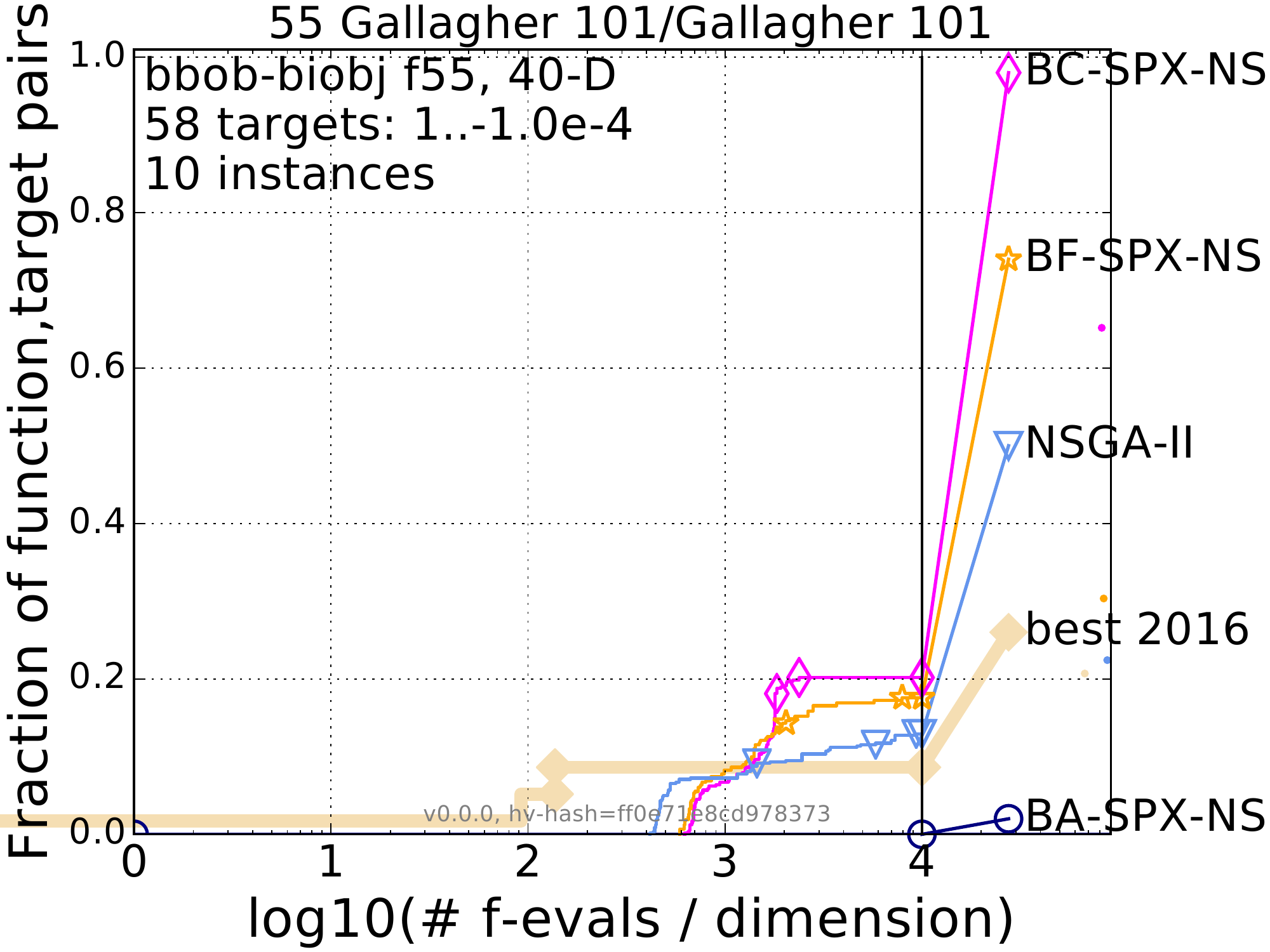}
    }
    \caption{
\small
Results of NSGA-II, BA-SPX-NS, BF-SPX-NS, and BC-SPX-NS on $\vector{f}_{54}$ and $\vector{f}_{55}$ with $n=40$.
%Results of three EMOAs with SPX and the ranking method in NSGA-II.
%The figures (a) and (b) show results which BC/BF outperforms BF/BC.
}
\label{fig:spx_nsgaii_each}
  \end{center}
\end{figure}

Note that BC-SPX-NS is not always the best optimizer on all 55 BBOB problems with $n=40$.
Figure \ref{fig:spx_nsgaii_each} shows results on $\vector{f}_{54}$ and $\vector{f}_{55}$ with $n=40$.
While BF-SPX-NS outperforms BC-SPX-NS on $\vector{f}_{54}$, BC-SPX-NS outperforms BF-SPX-NS on $\vector{f}_{55}$.
Similar to Figure \ref{fig:spx_nsgaii_each}, the best optimizer is different depending on the test problem.
We attempted to clarify which problem groups BC performs the best (e.g., BC has the best performance on multimodal problems with weak global structure such as $\vector{f}_{54}$ and $\vector{f}_{55}$).
Unfortunately, we could not find such a result.
An in-depth analysis is needed to understand on which problems BC performs well or poorly.

% (e.g., BC has a certain advantage compared to BF on multi-modal problems).
%% Finding a problem class which BC outperforms BF is an avenue for future work.

%
%% In \cite{Akimoto10}, Akimoto analyzes the performance of elitist and non-elitist GAs for single-objective continuous optimization.
%% The results in \cite{Akimoto10} shows that the non-elitist GAs has good performance on multi-modal problems.
%% To obtain the  , we carefully examined results on all 55 test problems.
%% Unfortunately, we could not find a common findings (e.g., BC has a certain advantage compared to BF on multi-modal problems).
%% Finding a problem class which BC outperforms BF is an avenue for future work.
%As shown in Figure \ref{fig:spx_nsgaii_each} (a), BF have a certain advantage compared to BC for some test problems.

% does not work well Why do adaptive DEs fail?
\subsection{Why does BA perform poorly?}
\label{sec:analysis_ba}

%% Figure \ref{fig:emoa_spx_nsgaii} shows both BF and BC perform significantly better than the original NSGA-II for all $n \in \{2, 3, 5, 10, 20, 40\}$.
%% It is interesting to note that BC is outperformed by BF, but outperforms NSGA-II.
%% This mean that an EMOA with non-elitist update method can perform better than NSGA-II when using SPX.
%% Among the two update method, BF outperforms BC for $n \leq 20$.
%% However, BC outperforms BF for $N=40$.

%% Figure \ref{fig:emoa_spx_nsgaii} shows BA-SPX performs better than BF-SPX, BC-SPX, and NSGA-II within $1\,000 \times n$ function evaluations for $n=2$.

%In Subsection \ref{sec:comparison_three_selections}, BA-SPX-NS performs poorly on problems with many decision variables.

%We also discuss the superior performance of BF-SPX-NS and BC-SPX-NS.

%% Figure 
%% This good performance of BA-SPX in the early stage of the search is due to the replacement without restriction.
%% However, as the $n$ increased, the performance of BA-SPX significantly degrades.
%% Figure \ref{} shows a distribution of children and the population of BA-SPX on $f_1$ with $n=40$.
% If the ranks of the children are better than those of other $\mu-k$ individuals in the population, most of them can enter the population.

Here, we discuss the poor performance of BA-SPX-NS observed in Subsection \ref{sec:comparison_three_selections}.
The biased distribution of children is likely to cause the poor performance of BA-SPX-NS.
As shown in Figure \ref{fig:dist_children} (d), SPX generates $\lambda$ children inside a simplex formed by $k$ parents.
If the $k$ parents are close to each other in the solution space, their $\lambda$ children are likely to be in local area.
If non-parents in the population are ranked worse than the children, the non-parents are replaced with the children in BA.
This means that non-parents in not-well-explored area cannot survive to the next iteration.
Thus, BA-SPX-NS is likely to lose diversity in the solution and objective spaces as the search progresses.
%

%% A small $\lambda$ pre vent BA from the premature convergence, but it is not appropriate for exploit the current search area.
%% Although BA has an essential issue, BF and BC are fine due to their restricted selection mechanisms.

%% of the children are better than those
%% , most of them can enter the population.
%% This much replacement degrades  the diversity of the population as shown in Figure \ref{}.
%For this reason, BA does not work well in SPX.

\begin{figure}[t]
\newcommand{\widthvar}{0.23}
  \begin{center} 
\subfloat[$n=10$]{\includegraphics[width=\widthvar\textwidth]{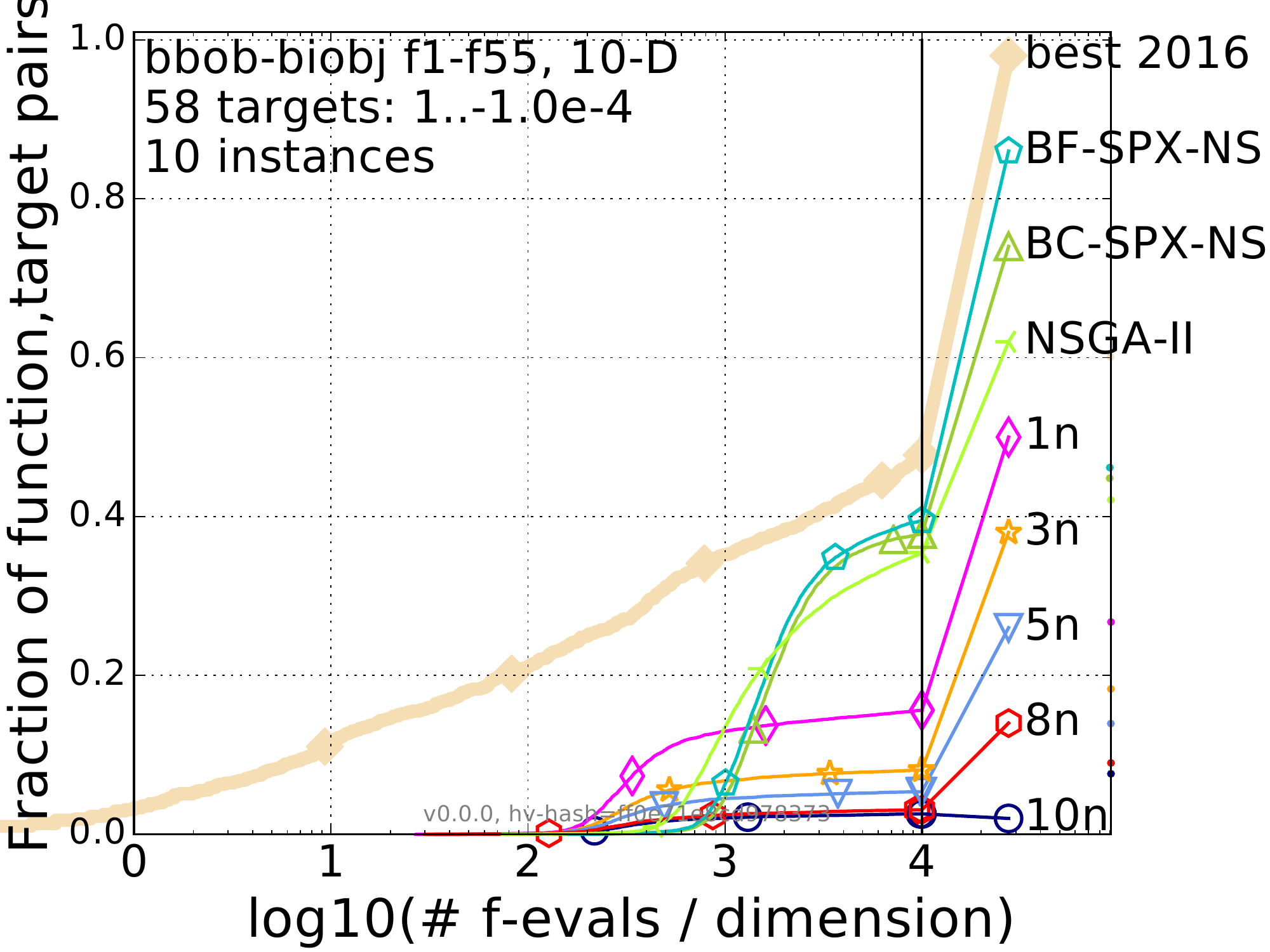}}
\subfloat[$n=40$]{\includegraphics[width=\widthvar\textwidth]{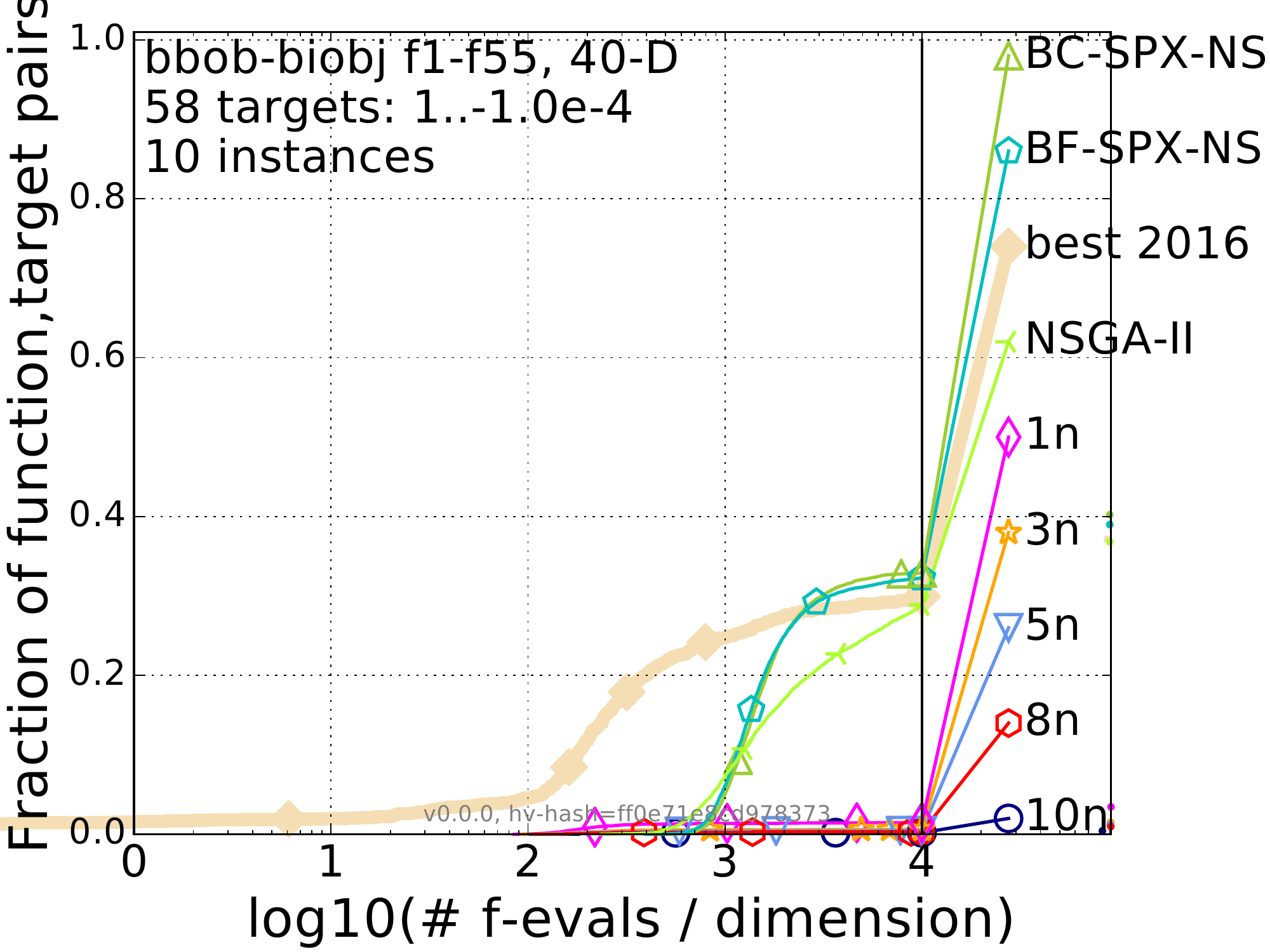}}
\caption{
  \small
  %
%Results of BA-SPX-NS with $\lambda \in \{1n, 3n, 5n, 8n, 10n\}$.
Results of BA-SPX-NS with various $\lambda$ values.
}
\label{fig:emoa_ba_various_lambda}
  \end{center}
\end{figure}

One may think that the above-mentioned issue caused by the biased distribution of children can be addressed by setting $\lambda$ to a small value.
Figure \ref{fig:emoa_ba_various_lambda} shows BA-SPX-NS with $\lambda \in \{1n, 3n, 5n, 8n, 10n\}$ on all 55 BBOB problems with $n \in \{10, 40\}$.
In Figure \ref{fig:emoa_ba_various_lambda}, ``$10n$'' is identical to BA-SPX-NS in Figure \ref{fig:emoa_spx_nsgaii}.
Figure \ref{fig:emoa_ba_various_lambda} also shows the results of NSGA-II, BF-SPX-NS and BC-SPX-NS derived from Figure \ref{fig:emoa_spx_nsgaii}.
Figure \ref{fig:emoa_ba_various_lambda} shows that the performance of BA-SPX-NS can be improved by setting $\lambda$ to a small value.
However, BA-SPX-NS with any $\lambda$ is outperformed by NSGA-II, BF-SPX-NS, and BC-SPX-NS at the later stage. % for $n \in \{10, 40\}$.

%Also, Figure \ref{fig:emoa_ba_various_lambda} (b) shows that BA-SPX-NS with any $\lambda$ performs significantly worse than the other three EMOAs for $n=40$.
%Thus, the results show that it is difficult to essentially improve the performance of BA-SPX-NS by adjusting $\lambda$.

%The reason why BA-SPX-NS with small $\lambda$ values do not work well is that

In general, a large enough number of children are necessary to find better solutions in the current search area \cite{Akimoto10}.
Thus, BA is in a dilemma.
A large $\lambda$ value is helpful for BA to exploit the current search area, but it causes premature convergence.
A small $\lambda$ value can prevent BA from the premature convergence, but it is not sufficiently large to exploit the current search area.
In addition to SPX, we observed the same issue in other crossover methods (except for SBX).

%The same issue can be found in other crossover methods (except for SBX).
%A similar results were found in 

%In contrast to BA, BF and BC performs well

In contrast to BA, only $k$ parents can be replaced with children in BF and BC.
This restricted replacement in BF and BC can help the population to maintain the diversity.
Even if non-parents in not-well-explored area are dominated by the children, the non-parents can survive to the next iteration with no comparison.
Thus, BF and BC can address the BA's dilemma.
In fact, BF-SPX-NS and BC-SPX-NS perform significantly better than BA-SPX-NS.

%Thus, the restricted selection in BF and BC can help the population to maintain the diversity.
%Although BA has an essential issue, BF and BC are fine due to their restricted selection mechanisms.

%% He struggled with a dilemma
%% there is a dilemma in $\lambda$.

%% For results for $n=5$, small $\lambda$ values lead high performance of BA.
%% Thus, a parameter tuning of $\lambda$ is beneficial for low-dimensional problems.
%% However,

\subsection{Advantages and disadvantages of BC}
\label{sec:advantage_bc}

%The results 
As shown in Subsection \ref{sec:comparison_three_selections}, the non-elitist BC performs better than the elitist BF for $n=40$.
Here, we discuss the advantages and disadvantages of BC compared with BF.

\begin{figure}[t]
\newcommand{\widthvar}{0.235}
  \begin{center} 
    \subfloat[$I_{\rm COCO}$ values]{\includegraphics[width=\widthvar\textwidth]{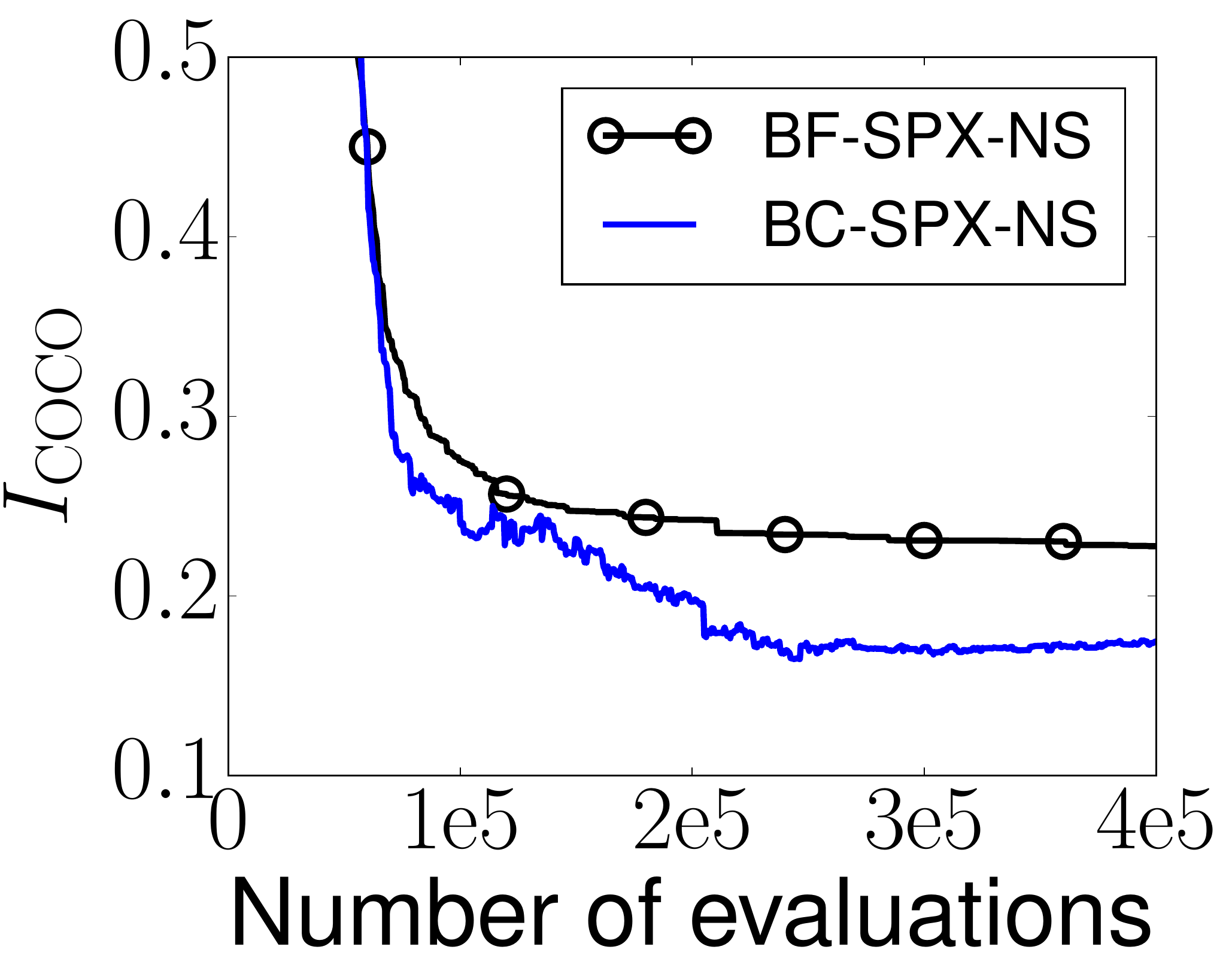}}
    \subfloat[Num. replacements]{\includegraphics[width=\widthvar\textwidth]{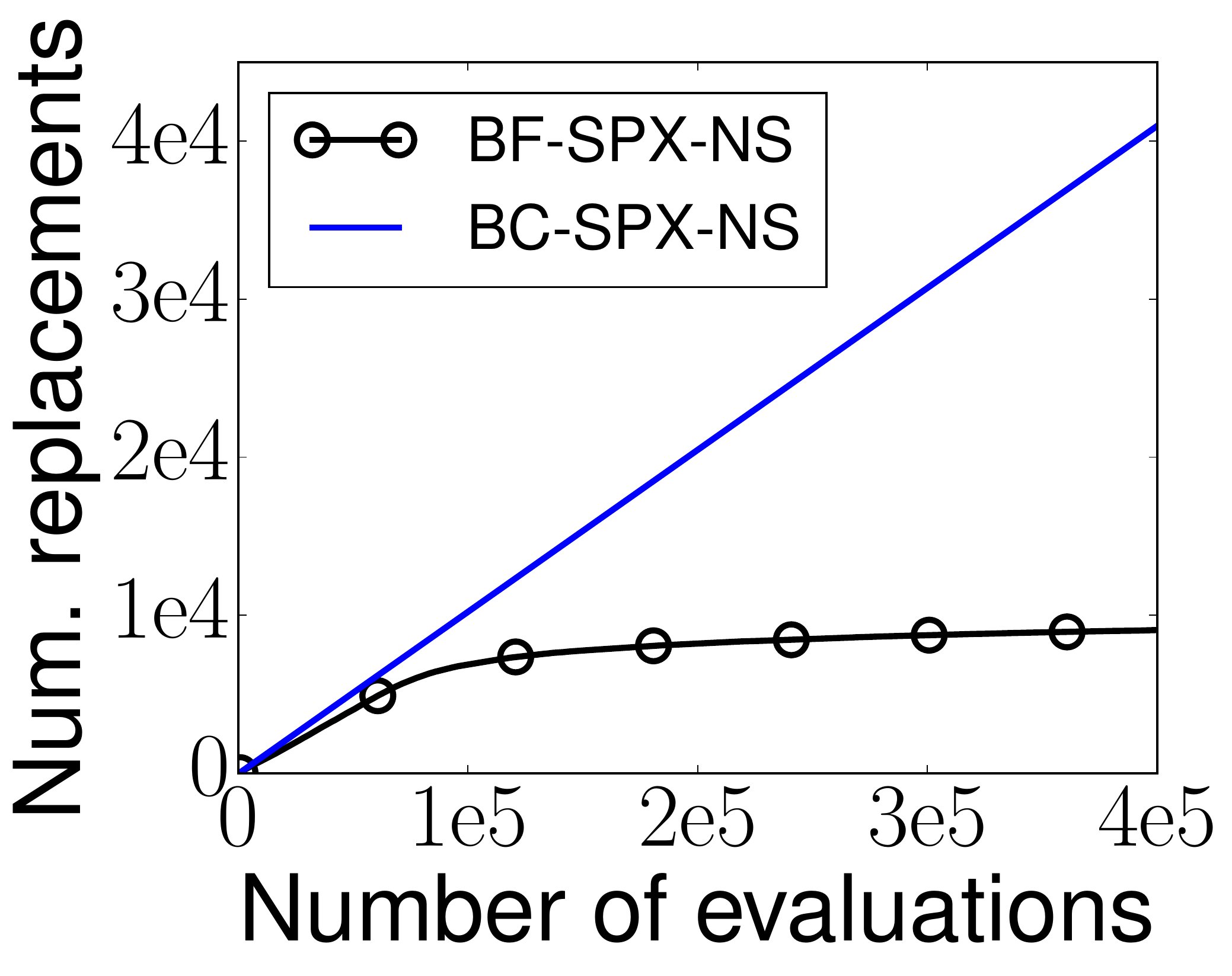}}    
%    \subfloat[BC]{\includegraphics[width=\widthvar\textwidth]{graph/2d_obj/bc_spx_nsgaii_f55_d40.pdf}}
    \caption{
\small
(a) Raw $I_{\rm COCO}$ indicator values on $\vector{f}_{46}$ with $n=40$ (lower is better). (b) Cumulative number of parents replaced by children.
Results of a single run are shown.
%(a) COCO's performance indicator ($I_{\rm COCO}$) values on $f_{46}$ (Rastrigin/Rastrigin) with $n=40$. (b) Cumulative number of replacements by children.
%Distribution of non-dominated solutions found by BF and BC on $f_{55}$ with $d=40$.
}
\label{fig:rast_rast}
  \end{center}
\end{figure}

Figure \ref{fig:rast_rast} (a) shows raw $I_{\rm COCO}$ indicator values of {\em the population} in BF-SPX-NS and BC-SPX-NS on $\vector{f}_{46}$ with $n=40$, which consists of two rotated Rastrigin function instances.
In all 55 BBOB test problems, $\vector{f}_{46}$ can be viewed as being a representative multimodal problem.
We slightly modified the COCO software to calculate the $I_{\rm COCO}$ value of the population (not the external archive).
A lower raw $I_{\rm COCO}$ value is better.
The range of the $I_{\rm COCO}$ value in Figure \ref{fig:rast_rast} (a) is limited to $[0.1, 0.5]$ in order to focus on the interesting behavior of BC-SPX-NS.
Although the $I_{\rm COCO}$ value of the elitist BF-SPX-NS almost\footnote{The monotonic improvement of the hypervolume value over time is guaranteed only when using the unbounded external archive \cite{Lopez-IbanezKL11}.} monotonically decreases as the search progresses, that of the non-elitist BC-SPX-NS is unstable.
Since BC does not maintain best-so-far non-dominated solutions in the population, its $I_{\rm COCO}$ value sometimes deteriorates compared with the previous iteration.

Figure \ref{fig:rast_rast} (b) shows the cumulative number $c$ of parents replaced by children.
In BF-SPX-NS, the evolution of $c$ clearly stagnates after $10^5$ function evaluations.
This result means that BF-SPX-NS rarely generates better children than parents.
In fact, the raw $I_{\rm COCO}$ value of BF-SPX-NS is not significantly improved after $10^5$ function evaluations, as shown in Figure \ref{fig:rast_rast} (a).
Since BC-SPX-NS always replaces $k$ parents with the best $k$ out of $\lambda$ children for every iteration, $c$ linearly increases.
Thus, the replacement of individuals in BC occurs more frequently than that in BF.
This property of BC is helpful for exploration of the search space.

%\cite{KirkpatrickGV83}.

%It is expected that 
%" algorithm is stuck in"
The above observations indicate that BC has a similar advantage to simulated annealing \cite{KirkpatrickGV83}, which can move to a worse search point.
As pointed out by Deb and Goel \cite{DebG01}, if an elitist EMOA prematurely converges to local Pareto optimal solutions, it is very likely to stagnate.
Unless the elitist EMOA finds better solutions far from the current search area, it cannot escape from local Pareto optimal solutions.
In contrast, the non-elitist BC always replaces $k$ parents with children regardless of the quality of $k$ parents.
While most elitist environmental selections accept only ``downhill'' moves on minimization problems,  the non-elitist BC can accept ``uphill'' moves as in simulated annealing.
The uphill moves in BC help the population to escape from local Pareto optimal solutions on {\em some} multimodal problems.
%As shown in Figure \ref{fig:rast_rast} (a), 

%However, the population in the non-elitist BC can move to worse .
%Since the population in BC can move to accepts the deteriorated solution

%BC prevents the population from stagnating local optima by allowing the worse improvement.
%However, this non-elisits selection prevents the population from stagnating local optima similar to the working principle of simulated annealing \cite{KirkpatrickGV83}.

However, BC has at least two disadvantages compared with the elitist BF.
First, as discussed in Subsection \ref{sec:comparison_three_selections}, BC performs worse than BF on some problems even with $n=40$.
Second, as reported in Subsection \ref{sec:comparison_three_selections}, BC performs worse than BF at the early stage.
Since BC can accept ``uphill'' moves as in simulated annealing, the exploitative ability of BC is worse than that of BF.
A deterministic or adaptive method of switching BC and BF may be promising to exploit their advantages.

%% \subsection{Comparison of BF and BC with the original NSGA-II}
%% %\label{sec:experimental_results}

\begin{figure*}[t]
\newcommand{\widthvar}{0.235}  
  \begin{center} 
    %
    %% \subfloat[AGR]{\includegraphics[width=\widthvar\textwidth]{graph/tmp/pprldmany_05D_noiselessall.pdf}}
    \subfloat[SBX]{
      \includegraphics[width=\widthvar\textwidth]{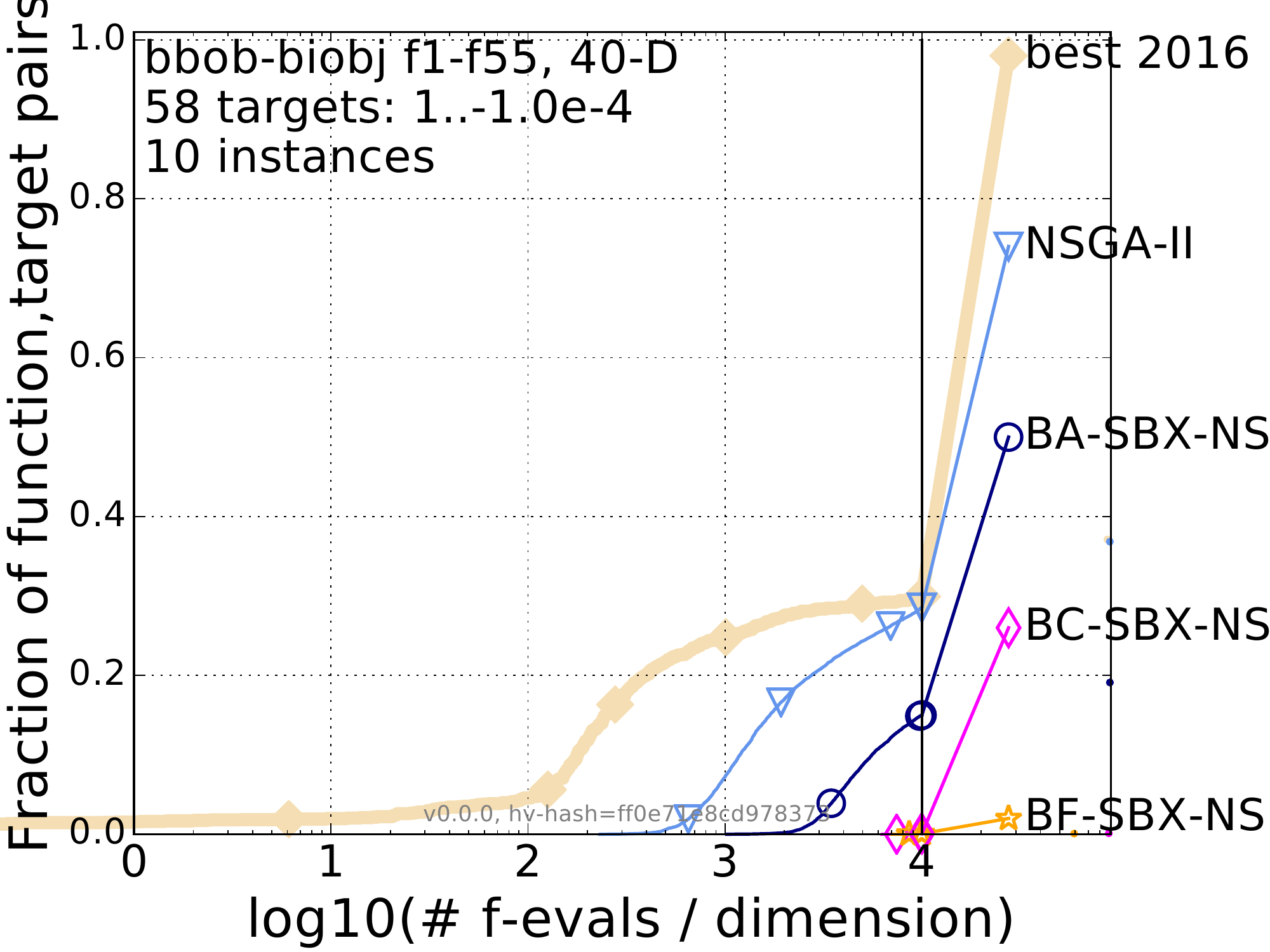}
    }
    \subfloat[BLX]{
      \includegraphics[width=\widthvar\textwidth]{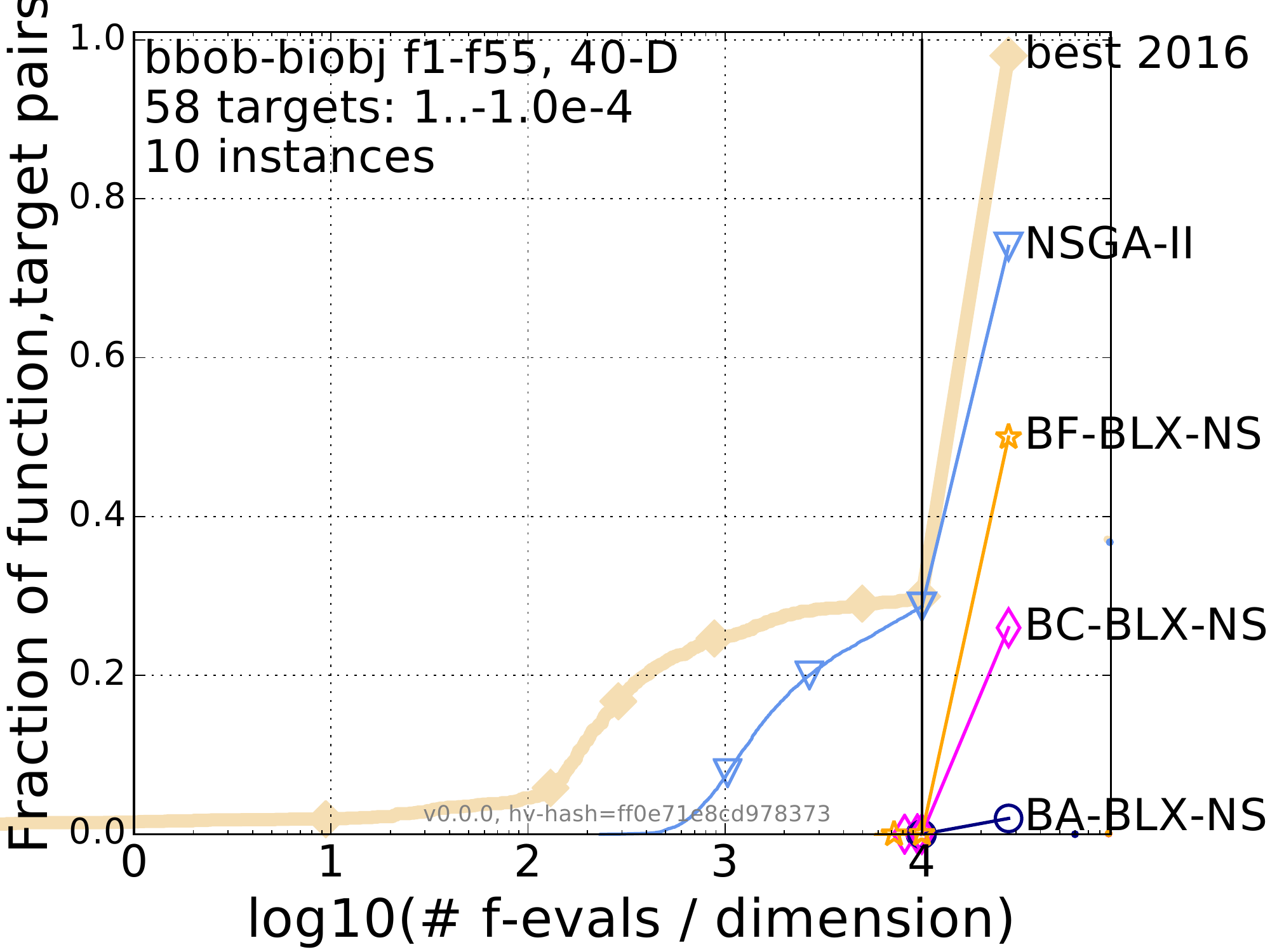}
    }
    \subfloat[PCX]{
      \includegraphics[width=\widthvar\textwidth]{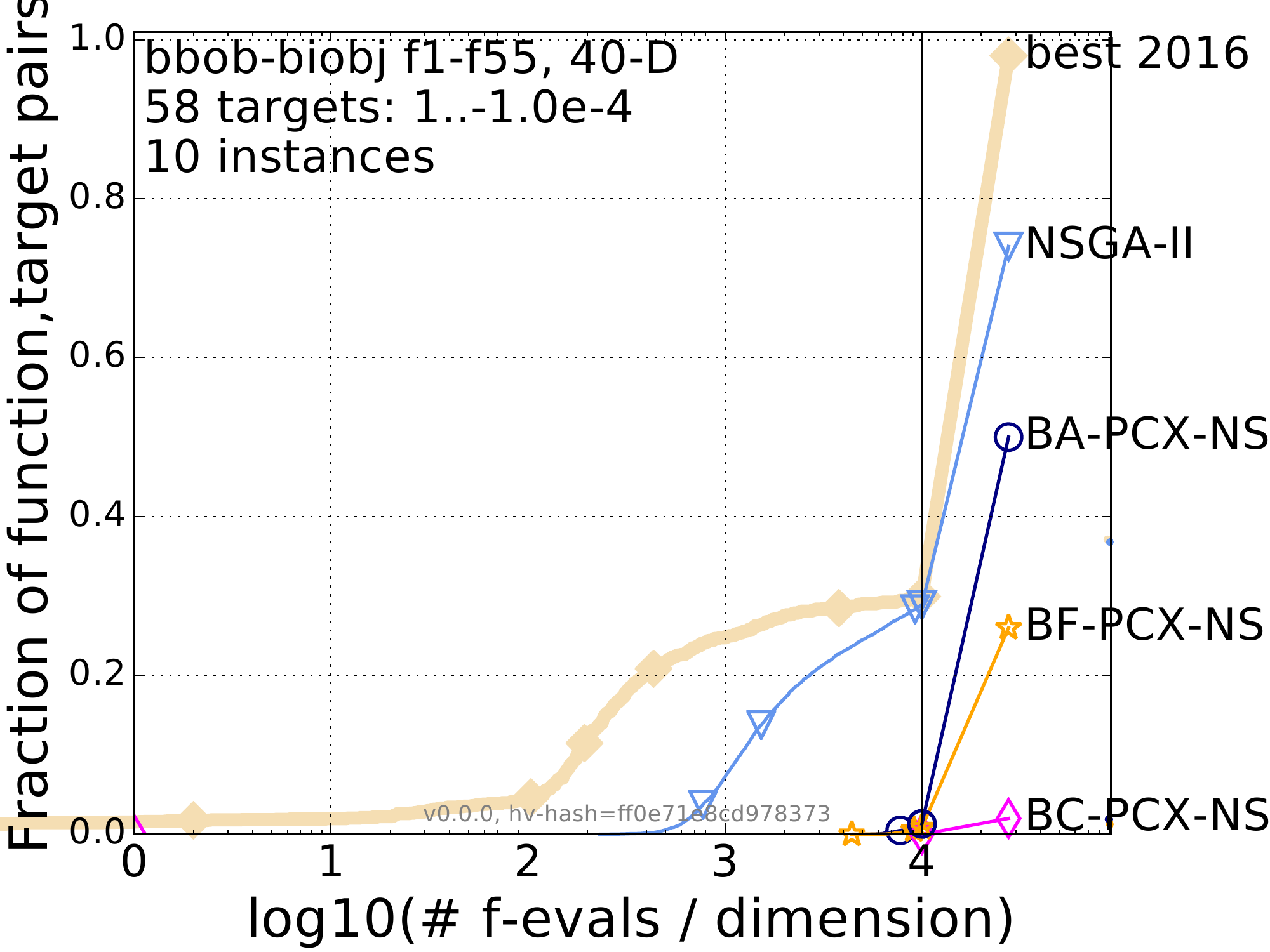}
    }
    \subfloat[REX]{
      \includegraphics[width=\widthvar\textwidth]{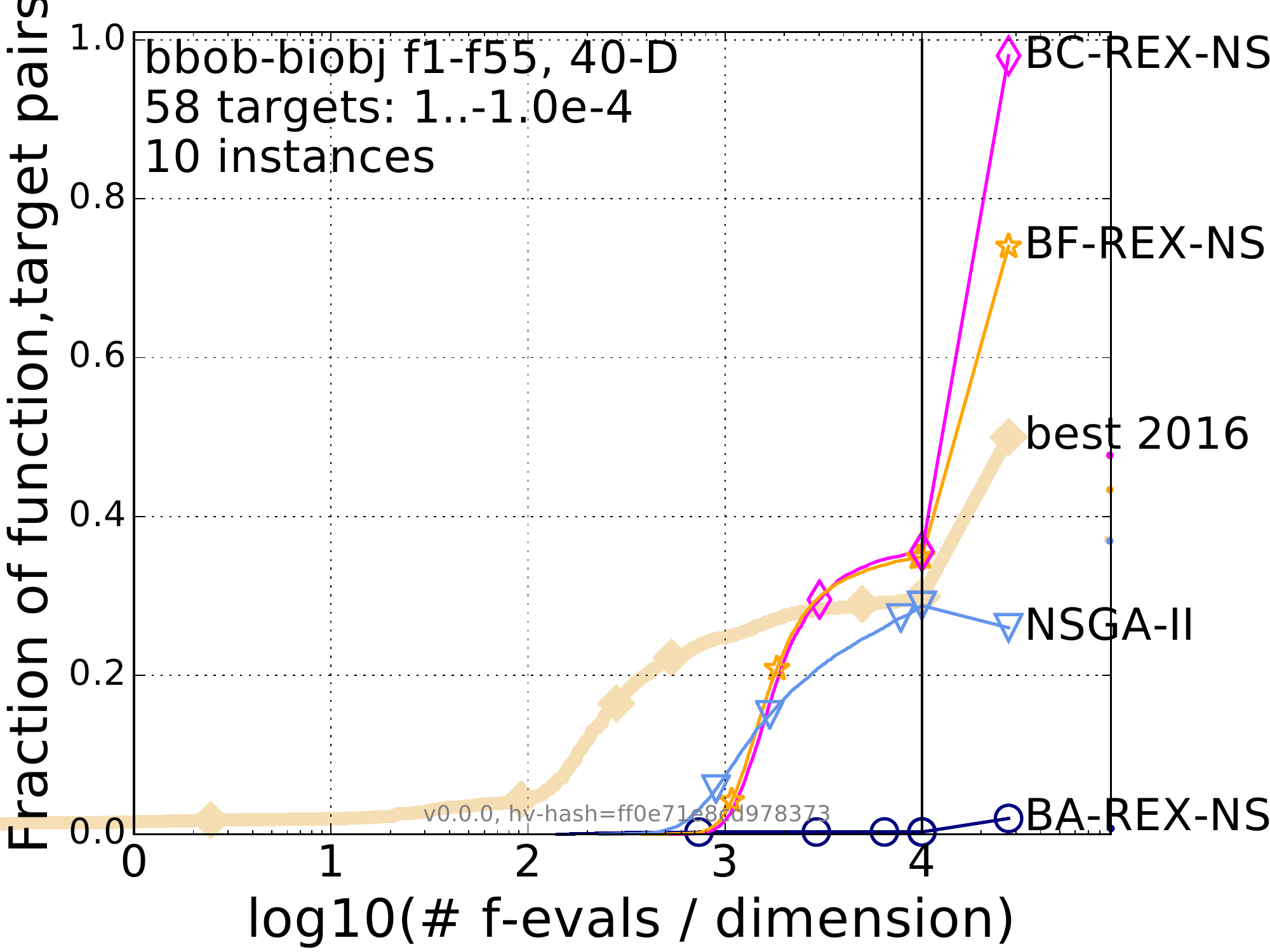}
    }
\caption{
  \small
Results of BA, BF, and BC with (a) SBX, (b) BLX, (c) PCX, and (d) REX on all 55 BBOB problems with $n=40$.
Results of the original NSGA-II are also shown.
}
\label{fig:emoa_threecross_nsgaii}
  \end{center}
\end{figure*}

\begin{figure*}[t]
\newcommand{\widthvar}{0.325}
  \begin{center} 
    %
    %% \subfloat[AGR]{\includegraphics[width=\widthvar\textwidth]{graph/tmp/pprldmany_05D_noiselessall.pdf}}
    \subfloat[SPEA2]{
      \includegraphics[width=\widthvar\textwidth]{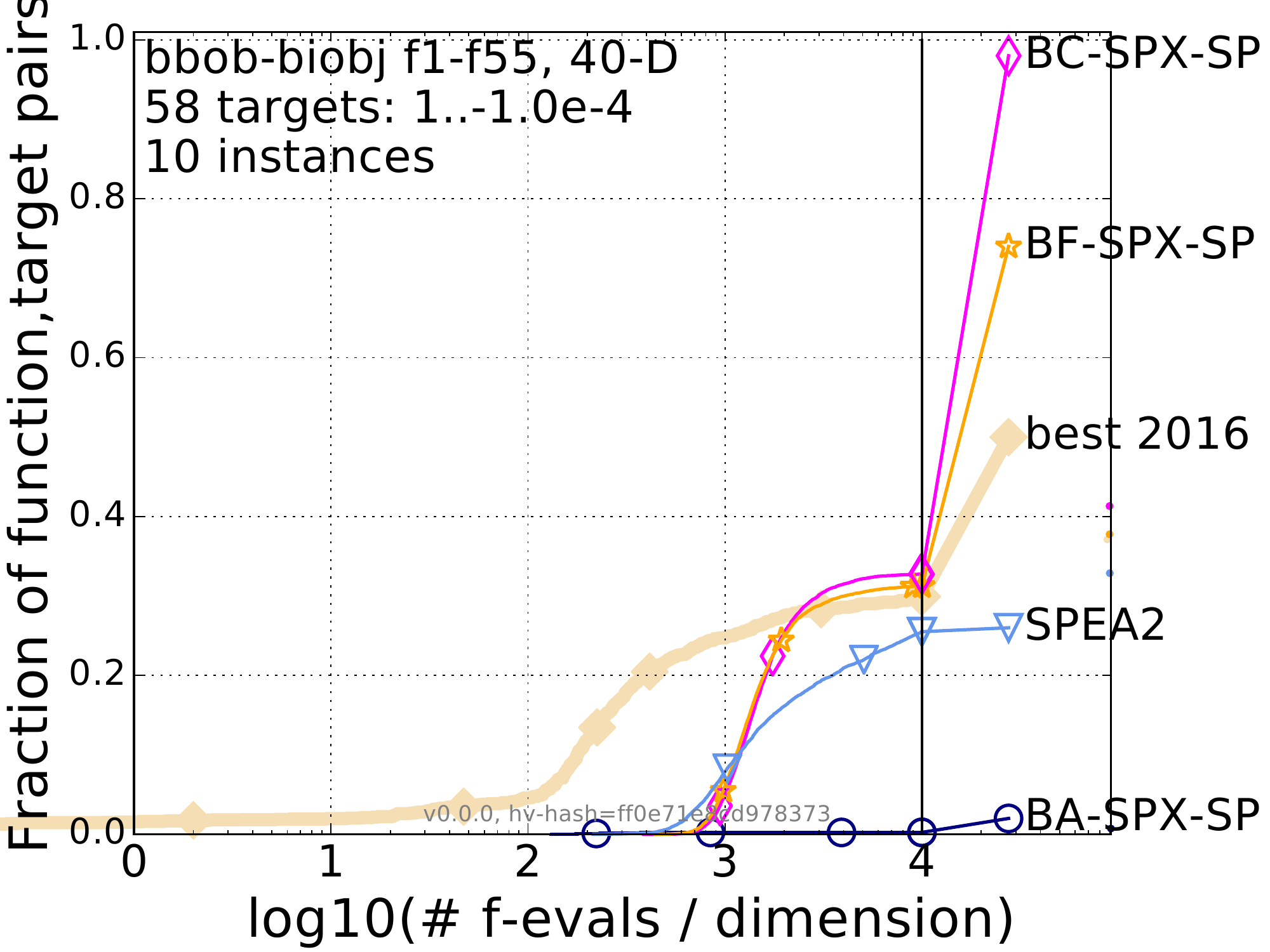}
    }
    \subfloat[SMS-EMOA]{
      \includegraphics[width=\widthvar\textwidth]{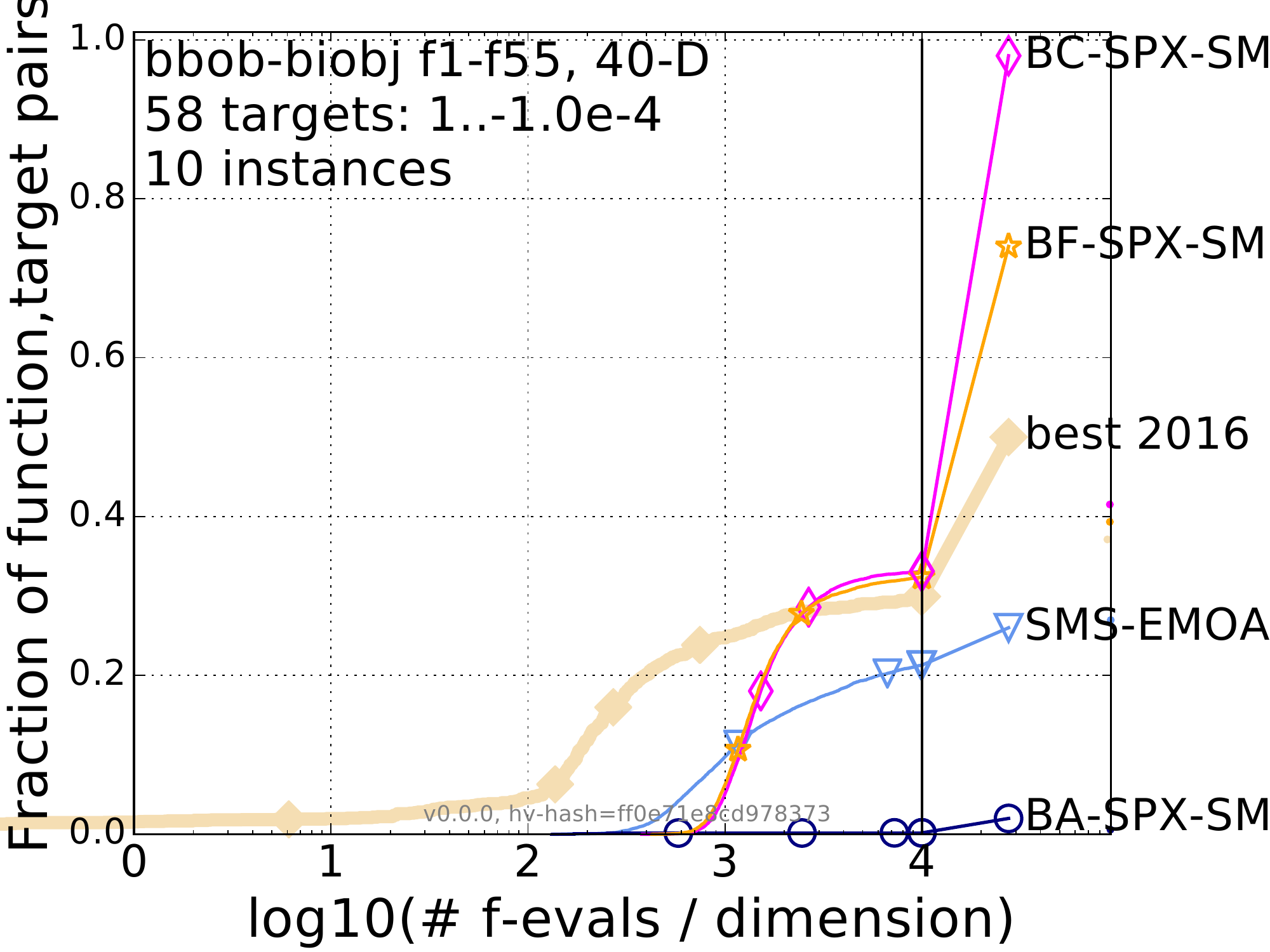}
    }
    \subfloat[IBEA]{
      \includegraphics[width=\widthvar\textwidth]{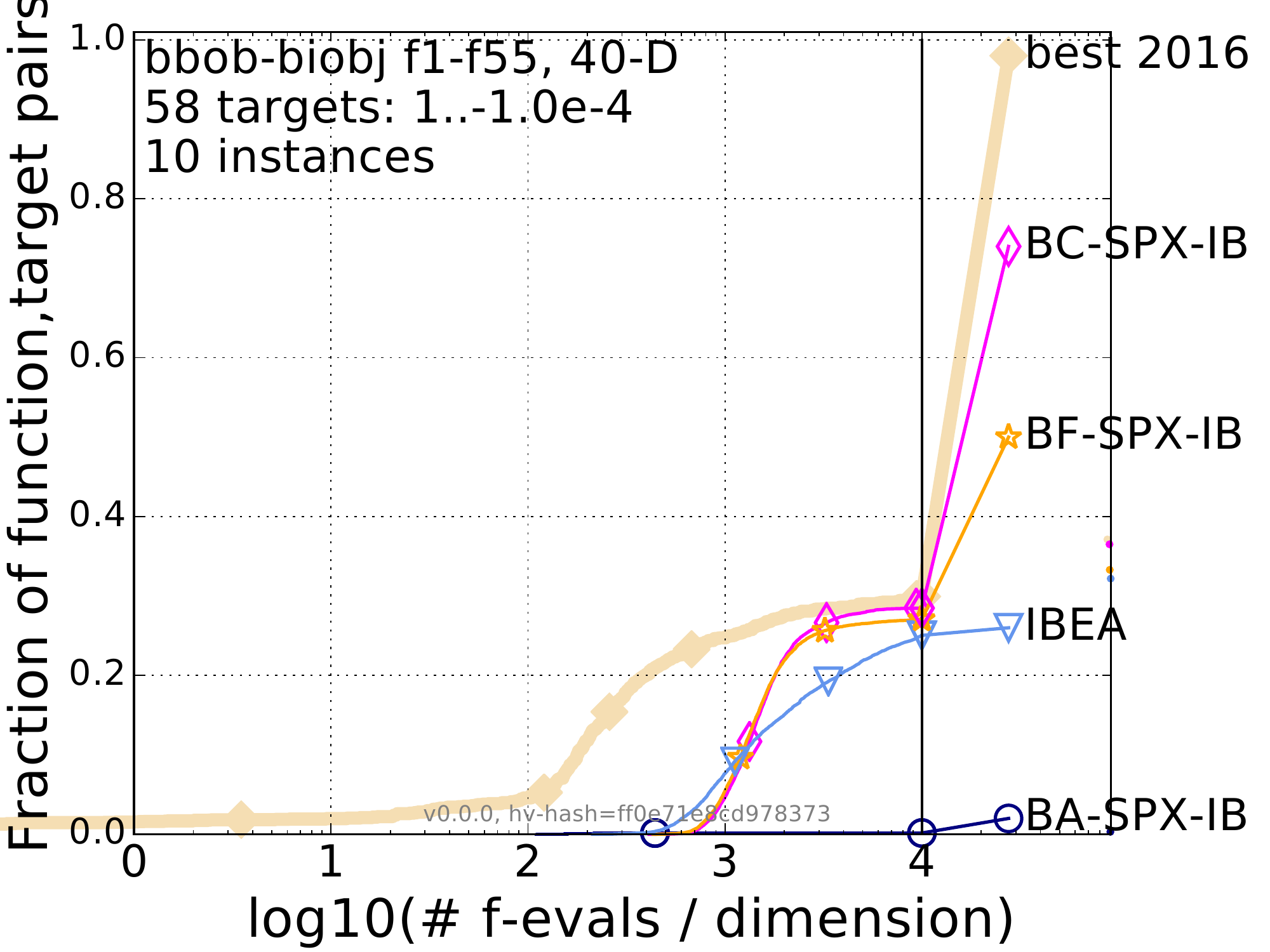}
    }
\caption{
  \small
Results of BA, BF, and BC with the ranking methods in (a) SPEA2, (b) SMS-EMOA, and (c) IBEA on all 55 BBOB problems with $n=40$.
Results of the original SPEA2, SMS-EMOA, and IBEA are also shown.
}
\label{fig:emoa_various_ranking}
  \end{center}
\end{figure*}

%\subsection{Comparison of BA, BF, and BC with other crossover methods}
\subsection{Which crossover methods are suitable for the non-elitist BC?}
\label{sec:other_crossovers}

The results in Subsection \ref{sec:comparison_three_selections} show that BC-SPX-NS outperforms BA-SPX-NS, BF-SPX-NS, and NSGA-II for $n=40$.
Here, we examine which crossover methods are suitable for BC.
%Here, we examine whether the same conclusion holds when using other crossover methods.
We are not interested in which crossover method is best.
Even though BC-SPX-NS outperforms BC-PCX-NS, it does not mean that SPX performs better than PCX.
It only means that SPX is more suitable for BC than PCX.

%Even though BC-SPX-NS outperforms BC-SBX-NS, it does not mean that SPX performs better than SBX.
%Such a result means that SPX is more suitable for the BC selection than PCX.
%This supriror performance 

Figure \ref{fig:emoa_threecross_nsgaii} shows results of the three selections with SBX, BLX, PCX, and REX on all 55 BBOB problems with $n=40$.
Due to space constraint, only results for $n=40$ are shown here.
The NS ranking method is used in BA, BC, and BF.
Figure \ref{fig:emoa_threecross_nsgaii} (a) shows that BA-SBX-NS outperforms BF-SBX-NS and BC-SBX-NS.
This good performance of BA-SBX-NS is inconsistent with the results in Subsection \ref{sec:comparison_three_selections}.
Since SBX can generate children far from their parents as shown in Figure \ref{fig:dist_children}, the distribution of children discussed in Subsection \ref{sec:analysis_ba} does not significantly influence the performance of BA.
However, BA-SBX-NS performs worse than NSGA-II.
Figure \ref{fig:emoa_threecross_nsgaii} (b) and (c) show similar results.
The evolution of the three selections with BLX and PCX clearly stagnates. %at the early stage.
Figure \ref{fig:emoa_threecross_nsgaii} (d) shows that results with REX are consistent with the results with SPX.
BC-REX-NS is the best optimizer at the later stage.
BF-REX-NS also performs better than NSGA-II.

%% BC and BF outperforms NSGA-II and BA.
%% BC performs better than BF at the later stage.

In summary, SPX and REX are suitable for BC and BF, while SBX, BLX, and PCX are not suitable for them.
%BC-SPX-NS and BC-REX-NS perform significantly better than the original NSGA-II for $n=40$.
%
These results indicate that crossover methods with the preservation of statistics are suitable for BC (and BF).
As shown in Table \ref{tab:cross_properties}, only SPX and REX satisfy the preservation of statistics among the five crossover methods.
The results presented in \cite{Akimoto10} show that SPX and UNDX-$n$ (a special version of REX) are suitable for JGG (a similar selection to BC) in GAs for single-objective continuous optimization.
Interestingly, our results on continuous MOPs are consistent with the results on single-objective continuous optimization problems. 
A similarity analysis between single-objective optimizers and multi-objective optimizers as in \cite{WessingPBR17} may be interesting.

%An analysis between

%% real-coded 

%The results In \cite{Akimoto10}

%% BLX is a mean-centric crossover similar to SPX and REX, and PCX is rotationally invariant.
%% However, BLX and PCX do not satisfy the preservation of statistics.

%% In summary, BC with SPX and REX show the better performance than BA, BF, and the original NSGA-II for $m=40$.
%% Based on the results, crossover methods with the preservation of statistics are suitable for the non-elitist BC (and the elitist BF).
%% As shown in Table \ref{tab:cross_properties}, only SPX and REX satisfy the preservation of statistics.
%% BLX is a mean-centric crossover similar to SPX and REX, and PCX is rotationally invariant.
%% However, BLX and PCX do not satisfy the preservation of statistics.

%Among the four crossover methods investigeted here, 
%Figure \ref{fig:emoa_threecross_nsgaii} (d) 
%% On the other hand, results with REX shows similar results to SPX.
%% This is because...

\subsection{Comparison of BA, BF, and BC with other ranking methods}
%\subsection{RoComparison of BA, BF, and BC with other ranking methods}
\label{sec:other_enviromental_selections}

%The results presented in Subsection \ref{sec:comparison_three_selections} show that BC performs better than BF and BA for $n=40$ when using the ranking method in NSGA-II.

%The results presented in Subsection \ref{sec:comparison_three_selections} show that BC performs better than BF and BA for $n=40$ when using the ranking method in NSGA-II.

We used the NS ranking method in Subsection \ref{sec:comparison_three_selections}.
We investigate whether similar results can be obtained when using the SP, SM, and IB ranking methods (see Subsection \ref{sec:env_selection}).

Figure \ref{fig:emoa_various_ranking} shows the comparison of BA, BF, and BC with SP, SM, and IB for $n=40$.
We do not show results for $n \in \{2, 3, 5, 10, 20\}$, but they are similar to the results in Subsection \ref{sec:comparison_three_selections}.
SPX is used as a crossover method.
Figures \ref{fig:emoa_various_ranking} (a), (b), and (c) also show results of the original SPEA2, SMS-EMOA, and IBEA, respectively.

Figure \ref{fig:emoa_various_ranking} shows that results with SP, SM, and IB are consistent with the results with NS.
BF and BC outperform the original SPEA2, SMS-EMOA, and IBEA at the later stage.
BC is the best optimizer at the later stage.
The poor performance of BA can be observed in Figure \ref{fig:emoa_various_ranking}.
Our results show that the relative performance of BA, BF, and BC does not significantly depend on the choice of a ranking method.

%Interestingly, the original SMS-EMOA performs worse than the original IBEA, but BC-SPX-SM performs better than BC-SPX-IB.
%This result indicate that the restricted selections (BC and BF) and SPX can improve the perofromance of the original EMOAs.
%Suitable ranking methods for BC do not 
%Altough the performance of BF and BC with the ranking method in IBEA is worse than that with the ranking method in NSGA-II, SPEA2, and SMS-EMOA, all results are relatively

%Figure \ref{fig:emoa_various_ranking} shows that similar results to the NSGA-II are obtained using other ranking methods.

%% In the ranking methods of SPEA2 and IBEA, individuals are ranked based on their fitness values.
%% In IBEA, the additive $\epsilon$ indicator is used.
%% In SMS-EMOA, the individuals are grouped according to  their non-domination levels similar to NSGA-II.
%% Then, ties are broken by the hypervolume loss.

%% file: conclusion.tex
\section{Conclusion}
\label{sec:conclusion}

%After SPEA has been proposed in 1999, it has been consided in the EMO community that eliteist EMOAs must outperformn non-elitist EMOAs.
%We demonstrated a counter example for this conventional wisdom.
% We used the five crossover methods (SBX, BLX, PCX, SPX, and REX) and the four ranking methods (NS, SP, SM, and IB). %in NSGA-II, SPEA2, SMS-EMOA, and IBEA.

%We revisited the perofrmance of non-eliteist EMOAs. %in NSGA-II, SPEA2, SMS-EMOA, and IBEA.

We examined the effectiveness of the two elitist selections (BA and BF) and the non-elitist selection (BC) on the bi-objective BBOB problem suite.
We used five crossover methods and four ranking methods. 
For about two decades, it has been considered that elitist EMOAs always outperform non-elitist EMOAs. 
Interestingly, our results show that the non-elitist BC performs better than the two elitist selections and the four original EMOAs (NSGA-II, SPEA2, SMS-EMOA, and IBEA) on the bi-objective BBOB problems with many decision variables when using the unbounded external archive and a crossover method with the preservation of statistics (i.e., SPX and REX).
The choice of a ranking method does not significantly influence the relative performance of BC.
We also analyzed the advantages and disadvantages of the non-elitist BC selection.

%We analyzed the advantages and disadvantages of the restricted replacement in BF and BC.

%We also analyzed BA, BF, and BC.

%
%
%Our results show that the relative performance of BA, BF, and BC does not significantly depend on 

%Interestingly, our results show that the non-elitist BC performs better than the elitist BA and BF on problems with many decision variables when using the unbounded external archive and a crossover method with the preservation of statistics (i.e., SPX and REX).

%% preforms significantly better than BF on 40-dimensional problems when using SPX and REX.
%% We also analyzed the behavior of  BA, BF, and BC.

%% %While an EMOA with BA or BF is elitist EMOA, an EMOA with BC is non-elitist EMOA.
%% We used the four crossover methods: SBX, PCX, SPX, and REX.
%% Our results show that BC preforms significantly better than BF on 40-dimensional problems when using SPX and REX.
%% We also analyzed the behavior of  BA, BF, and BC.

A number of interesting directions for future work remain.
Although only elitist EMOAs have been studied in the 2000s, our results indicate that efficient non-elitist EMOAs could be realized.
Designing non-elitist versions of MO-ES \cite{WessingPBR17} and MO-CMA-ES \cite{IgelHR07} based on BC may be promising.